\documentclass{IEEEtran}
\usepackage{graphicx}
\usepackage{subcaption}
\usepackage{multicol}
\usepackage{booktabs}
\usepackage{multirow}
\usepackage{gensymb}
\usepackage{adjustbox}
\usepackage{array}
\usepackage{makecell}
\usepackage{tikz}
\usepackage{url}
\usepackage[numbers,sort&compress]{natbib}

\usepackage[framemethod=tikz]{mdframed}
\usepackage{xcolor}

\definecolor{lightgray}{RGB}{200,200,200} 

\mdfdefinestyle{remarkstyle}{
    backgroundcolor=lightgray, 
    roundcorner=3pt,
    innerleftmargin=8pt, 
    innerrightmargin=8pt,
    innertopmargin=8pt,
    innerbottommargin=8pt,
    linewidth=0.5pt,
    linecolor=black!50 
}

\newcounter{insight}
\newenvironment{insight}{\refstepcounter{insight}
\vspace{0.2em} 
\begin{mdframed}[style=remarkstyle]
\noindent \textbf{Insight~\theinsight}: \em
}
{
\end{mdframed}
\vspace{0.2em} 
}

\begin{document}

\title{Towards an End-to-End (E2E) Adversarial Learning and Application in the Physical World}

\author{Dudi Biton$^{1}$, Jacob Shams$^{1}$, Satoru Koda$^{2}$, Asaf Shabtai$^{1}$, Yuval Elovici$^{1}$, Ben Nassi$^{1}$\\\

$^{1}$Ben-Gurion University of the Negev, $^{2}$Fujitsu Limited\\ 
\{bitondud, jacobsh, nassib\}@post.bgu.ac.il,
\{shabtaia, elovici\}@bgu.ac.il,
koda.satoru@fujitsu.com\\[0.5em]
\textbf{PAPLA Video Demonstration:} \url{https://www.youtube.com/watch?v=AtambR-sJD4}
}

\maketitle
\thispagestyle{empty}

\begin{abstract}
The traditional learning process of patch-based adversarial attacks, conducted in the digital domain and then applied in the physical domain (e.g., via printed stickers), may suffer from reduced performance due to adversarial patches' limited transferability from the digital domain to the physical domain. Given that previous studies have considered using projectors to apply adversarial attacks, we raise the following question: can adversarial learning (i.e., patch generation) be performed entirely in the physical domain with a projector? In this work, we propose the Physical-domain Adversarial Patch Learning Augmentation (PAPLA) framework, a novel end-to-end (E2E) framework that converts adversarial learning from the digital domain to the physical domain using a projector.
We evaluate PAPLA across multiple scenarios, including controlled laboratory settings and realistic outdoor environments, demonstrating its ability to ensure attack success compared to conventional digital learning-physical application (DL-PA) methods. We also analyze the impact of environmental factors, such as projection surface color, projector strength, ambient light, distance, and angle of the target object relative to the camera, on the effectiveness of projected patches. Finally, we demonstrate the feasibility of the attack against a parked car and a stop sign in a real-world outdoor environment.
Our results show that under specific conditions, E2E adversarial learning in the physical domain eliminates the transferability issue and ensures evasion by object detectors. 
Finally, we provide insights into the challenges and opportunities of applying adversarial learning in the physical domain and explain where such an approach is more effective than using a sticker.

\end{abstract}

\section{Introduction}
\label{sec:introduction}






In recent years, object detectors have been integrated in various systems that obtain data from the physical domain, including security cameras \cite{ouardirhi2024enhancing}, license plate recognition systems \cite{lubna2021automatic}, autonomous vehicles \cite{juyal2021deep}, etc. 
However, these object detectors are vulnerable to adversarial attacks, where small, often imperceptible, image alterations can cause object detection systems to misclassify input. 
These misclassifications pose significant challenges to the reliability and safety of these systems.

\begin{figure}[ht]
    \centering
    \begin{subfigure}[b]{0.24\linewidth} 
        \includegraphics[width=\linewidth]{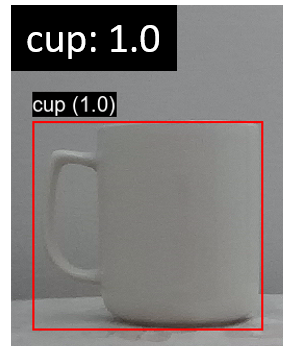}
        \caption{}
        \label{fig:intro_clean}
    \end{subfigure}
    \begin{subfigure}[b]{0.24\linewidth} 
        \includegraphics[width=\linewidth]{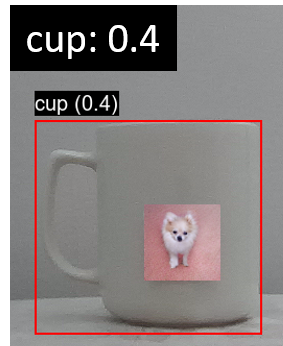}
        \caption{}
        \label{fig:intro_digital_digital}
    \end{subfigure}
    \begin{subfigure}[b]{0.24\linewidth} 
        \includegraphics[width=\linewidth]{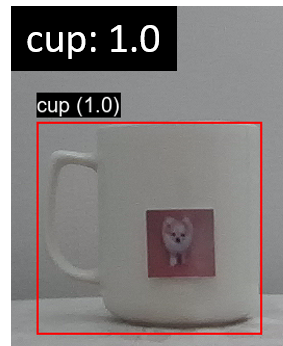}
        \caption{}
        \label{fig:intro_digital_physical}
    \end{subfigure}
    \begin{subfigure}[b]{0.24\linewidth} 
        \includegraphics[width=\linewidth]{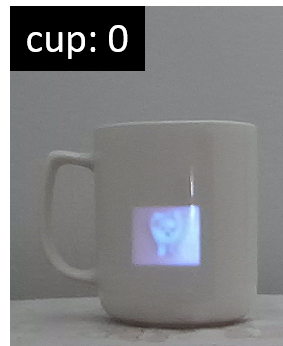}
        \caption{}
        \label{fig:intro_physical_physical}
    \end{subfigure}

    \caption{Application of adversarial patches in different learning scenarios. We applied the NAP~\cite{hu2021naturalistic} attack against the Faster R-CNN object detector in four different scenarios: (a) no application of NAP, (b) the adversarial patch was generated and applied to the object in the digital domain, (c) the patch was generated digitally and physically applied to the object as a sticker, and (d) using PAPLA, our E2E framework, the patch was generated and applied in the physical domain, causing the object detector to fail to detect the cup.}
    \label{fig:intro_visualization}
\end{figure}

Previous studies have carried out the generation process of adversarial examples in the digital domain \cite{chen2020hopskipjumpattack, guo2019simple,brendel2017decision,athalye2018synthesizingrobustadversarialexamples,song2018physical,lee2019physical,eykholt2018robust,katzav2025adversarialeak,chen2019shapeshifter,liu2018dpatch,zhang2018camou,thys2019fooling,xu2020adversarial,huang2020universal,wu2020making,zolfi2021translucent,hu2021naturalistic,jing2021too,tan2021legitimate,suryanto2022dta,hu2022adversarial,biton2023adversarial,jia2022fooling,huang2023t,zhu2023tpatch,hu2023physically,guesmi2024dap,weirevisiting, chengfull,zhu2022infrared,wei2023hotcold,wei2024infrared,zhu2024infrared,lovisotto2021slap,wen2024opticloak,hu2023adversarial,hwang2023adversarial,komkov2021advhat,lin2022real,wei2023unified}.
Using these digitally generated patches, several studies \cite{song2018physical, lee2019physical,eykholt2018robust,chen2019shapeshifter,liu2018dpatch,zhang2018camou,thys2019fooling,huang2020universal,wu2020making,hu2021naturalistic,tan2021legitimate,suryanto2022dta,weirevisiting,wei2023unified,wei2024infrared,zhu2024infrared} printed them as stickers and applied them in the physical domain.
However, these methods may face limitations due to the challenge of transferability between the digital and physical domains.
Specifically, patches that are effective in the digital domain may not perform as intended when applied in the physical domain.
Figure~\ref{fig:intro_visualization} illustrates this issue. In the non-adversarial case (Figure~\ref{fig:intro_clean}), the object detector identifies the cup with a high confidence score of 1.0. In the \textit{digital learning-digital application} (DL-DA) scenario (Figure~\ref{fig:intro_digital_digital}), the adversarial patch (applied digitally) successfully reduces the confidence score to 0.4, achieving its goal. However, when the same patch is printed and applied in the physical domain, as shown in the \textit{digital learning-physical application} (DL-PA) scenario (Figure~\ref{fig:intro_digital_physical}), it fails to affect the object detector, which maintains a confidence score of 1.0.
This issue of transferability between the digital and physical domains raises two essential questions; (1) Can adversarial patch learning be conducted entirely in the physical domain to ensure that physical patches maintain their effectiveness in the application phase? (2) Under what constraints may the physical learning approach yield better results than the traditional digital learning approach? 


In this paper, we address the transferability issue by extending digital adversarial learning methodologies to the physical domain. We start by adapting the learning process of digital patch attacks, in particular the Dpatch~\cite{liu2018dpatch} attack and the Naturalistic Adversarial Patch (NAP)~\cite{hu2021naturalistic} attack, for physical domain adversarial learning.
We extend the conventional approach, in which the adversarial learning process is carried out in a purely digital environment, by introducing Physical-domain Adversarial Patch Learning Augmentation (PAPLA), a new framework that enables the conversion of digital adversarial learning processes to their physical-domain equivalents. 
PAPLA aims to overcome the difficulty in transferring digital adversarial examples to physical-domain scenarios, by performing the adversarial patch learning process end-to-end (E2E) in the physical domain using a projector.
We find that transferring digital patches to the physical domain is affected by: (1) additional noise from external environmental factors and (2) difficulty in matching digital colors to printed colors.
By performing patch learning E2E in the physical domain, these factors are either avoided or integrated into the learning process.
As illustrated in Figure \ref{fig:intro_physical_physical}, PAPLA’s \textit{physical learning-physical application} (PL-PA) approach enables E2E adversarial patch learning and application directly in the physical domain, thus completely hiding the cup from the object detector.
Prior studies \cite{wen2024opticloak,hu2023adversarial,lovisotto2021slap,nassi2020phantom} have explored the use of projectors to apply adversarial attacks directly in the physical domain, leveraging light-based projections to mislead detection systems. We leverage projectors since they enable rapid iteration of adversarial patches in order to enable patch learning in the physical domain.

Our research demonstrates that PAPLA improves the confidence score reduction of adversarial patches in real-world settings, under specific conditions. By incorporating environmental factors like distance, angle, and lighting, into the learning process, PAPLA enhances robustness compared to traditional \textit{digital learning-physical application} methods.
However, its effectiveness is influenced by several factors.
For example, the color of the projection surface impacts performance, with lighter surfaces yielding better results. Furthermore, PAPLA introduces higher $L_{2}$ and $L_{\infty}$ norms compared to \textit{digital learning-physical application} patches, indicating a trade-off between image quality and the success of the attack. Environmental factors, including the distance and angle of the camera and the light intensity of the projector, also play a critical role; Greater projector strength and optimized camera positioning improve attack effectiveness.
While PAPLA demonstrates clear advantages in controlled environments, it requires careful setup. 
Most of the experiments described in this work were conducted in a controlled laboratory environment to ensure consistency and reproducibility. However, we also conduct experiments in an outdoor environment to evaluate PAPLA's robustness under realistic and dynamic conditions and demonstrate E2E physical learning and application against a parked car and a stop sign.

\textbf{Contributions.} Our contributions can be summarized as follows:
\textbf{(1)} In contrast to previous works where adversarial learning and attack application were conducted in different domains (e.g., \cite{athalye2018synthesizingrobustadversarialexamples,song2018physical,lee2019physical,eykholt2018robust,chen2019shapeshifter,liu2018dpatch,zhang2018camou,thys2019fooling,xu2020adversarial,huang2020universal,wu2020making,zolfi2021translucent,hu2021naturalistic,jing2021too,tan2021legitimate,suryanto2022dta,hu2022adversarial,jia2022fooling,huang2023t,zhu2023tpatch,hu2023physically,guesmi2024dap,weirevisiting, chengfull,zhu2022infrared,wei2023hotcold,wei2024infrared,zhu2024infrared,lovisotto2021slap,wen2024opticloak,hu2023adversarial,hwang2023adversarial,komkov2021advhat,lin2022real}), we take the first step in conducting E2E adversarial learning entirely in the physical domain. We present PAPLA, a framework that converts the adversarial learning of existing digital adversarial attacks to the physical domain. We convert two attacks (Dpatch~\cite{liu2018dpatch} and NAP~\cite{hu2021naturalistic}) from \textit{digital learning-digital application} and \textit{digital learning-physical application} scenarios to a \textit{physical learning-physical application} scenario and ensure the success of the attacks in the physical world.
\textbf{(2)} We perform a detailed analysis of the factors that influence the effectiveness of E2E physical domain adversarial attacks. Specifically, we investigate environmental factors including projector strength, ambient lighting, camera distance, camera angle, and projection surface, in order to determine their impact on attack effectiveness.
\textbf{(3)} We compare the results obtained from three distinct adversarial learning scenarios: \textit{digital learning-digital application} (patch learning and application are performed in the digital domain), \textit{digital learning-physical application} (patch learning is performed digitally, and printed/applied in the physical domain - the current practice for physical-domain attacks), and \textit{physical learning-physical application} (our framework - patch learning and application are performed in the physical domain). 

\textbf{Structure.} The rest of the paper is structured as follows: In Section \ref{sec:motivation} we present the motivation of E2E physical adversarial learning. This is followed by a detailed explanation of our threat model and method in Section \ref{sec: threat_model-method}.
We present a detailed analysis of PAPLA in Section \ref{sec:analysis}, and our evaluations in Section \ref{sec:evaluations}. We then discuss the limitations of E2E adversarial learning in the physical domain in Section \ref{sec:limitations}.
In Section \ref{sec:related-work} we review related work in the field.
We conclude with a summary of our findings and suggestions for future research in Section \ref{sec:discussion}.

\textbf{Ethical considerations.} The outdoor experiments in this study were approved by our university and conducted in its territory. 
We made sure to carry out the experiments at specific times and locations determined by the university to avoid affecting pedestrians and cars.
\section{Transferability Between Digital and Physical Domains}
\label{sec:motivation}






\begin{table*}[]
\centering
\caption{Comparison of generated adversarial patches and their effects on the Faster R-CNN and YOLOv3 object detectors across digital and physical domains.}
\renewcommand{\arraystretch}{1.2} 
\resizebox{\textwidth}{!}{%
\begin{tabular}{|>{\centering\arraybackslash}m{2.5cm}|>{\centering\arraybackslash}m{2.5cm}|>{\centering\arraybackslash}m{3.5cm}|>{\centering\arraybackslash}m{4cm}|>{\centering\arraybackslash}m{4cm}|>{\centering\arraybackslash}m{4cm}|>{\centering\arraybackslash}m{4cm}|>{\centering\arraybackslash}m{4cm}|>{\centering\arraybackslash}m{3cm}|}
\hline
\textbf{\LARGE Attack} & \textbf{\makecell{\LARGE Target \\ \LARGE Detector}} & \textbf{\LARGE Scenario} & \multicolumn{5}{c|}{\textbf{\LARGE Target Objects}} & \textbf{\makecell{\LARGE Avg. \\ \LARGE Conf. \\ \LARGE Score}} \\ \hline
\multirow{3}{*}{\LARGE DPatch} & \multirow{3}{*}{\LARGE YOLOv3} & \LARGE Non-Adversarial & 
\includegraphics[width=4.1cm, height=4.3cm]{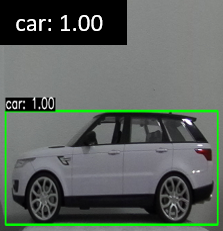} & 
\includegraphics[width=4.1cm, height=4.3cm]{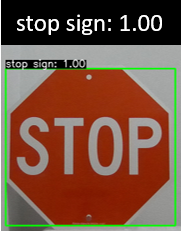} & 
\includegraphics[width=4.1cm, height=4.3cm]{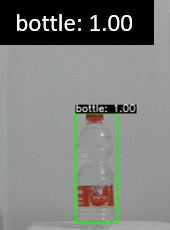} & 
\includegraphics[width=4.1cm, height=4.3cm]{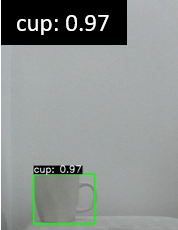} & 
\includegraphics[width=4.1cm, height=4.3cm]{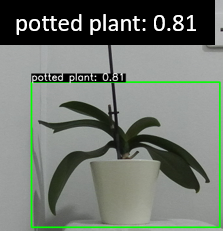} & \LARGE 0.96 \\ \cline{3-9} 
 &  & \LARGE Digital Learning - Digital Application & 
\includegraphics[width=4.1cm, height=4.3cm]{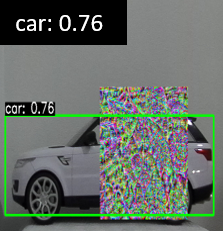} & 
\includegraphics[width=4.1cm, height=4.3cm]{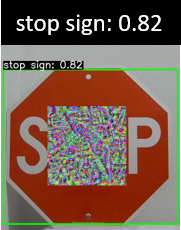} & 
\includegraphics[width=4.1cm, height=4.3cm]{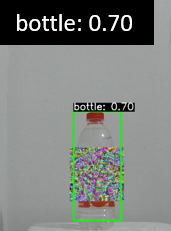} & 
\includegraphics[width=4.1cm, height=4.3cm]{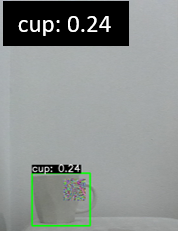} & 
\includegraphics[width=4.1cm, height=4.3cm]{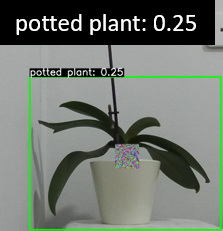} & \LARGE 0.55 \\ \cline{3-9} 
 &  & \LARGE Digital Learning - Physical Application & 
\includegraphics[width=4.1cm, height=4.3cm]{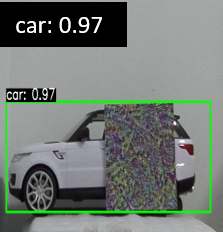} & 
\includegraphics[width=4.1cm, height=4.3cm]{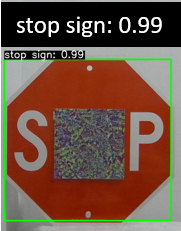} & 
\includegraphics[width=4.1cm, height=4.3cm]{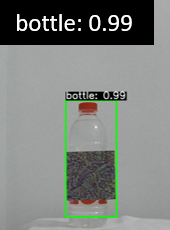} & 
\includegraphics[width=4.1cm, height=4.3cm]{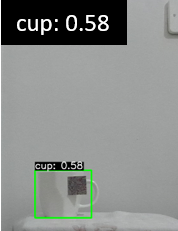} & 
\includegraphics[width=4.1cm, height=4.3cm]{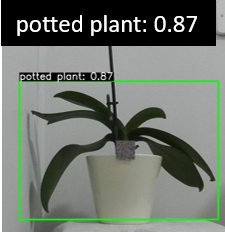} & \LARGE 0.88 \\ \hline
\multirow{3}{*}{\makecell{\LARGE Robust \\ \LARGE DPatch}} & \multirow{3}{*}{\makecell{\LARGE Faster \\ \LARGE R-CNN}} & \LARGE Non-Adversarial & 
\includegraphics[width=4.1cm, height=4.3cm]{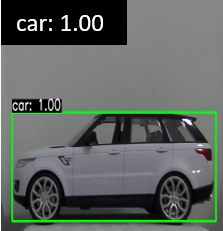} & 
\includegraphics[width=4.1cm, height=4.3cm]{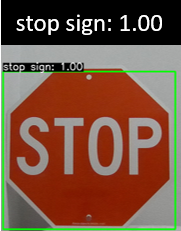} & 
\includegraphics[width=4.1cm, height=4.3cm]{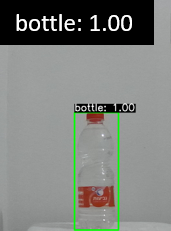} & 
\includegraphics[width=4.1cm, height=4.3cm]{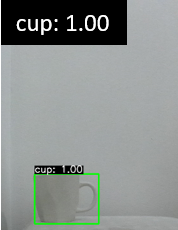} & 
\includegraphics[width=4.1cm, height=4.3cm]{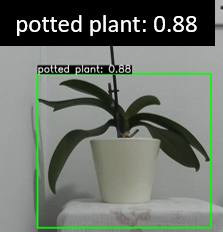} & \LARGE 0.98 \\ \cline{3-9} 
 &  & \LARGE Digital Learning - Digital Application & 
\includegraphics[width=4.1cm, height=4.3cm]{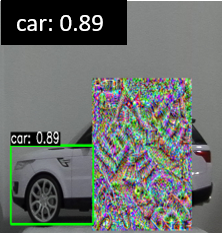} & 
\includegraphics[width=4.1cm, height=4.3cm]{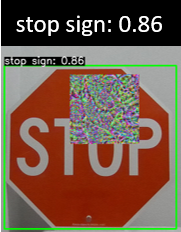} & 
\includegraphics[width=4.1cm, height=4.3cm]{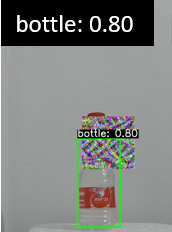} & 
\includegraphics[width=4.1cm, height=4.3cm]{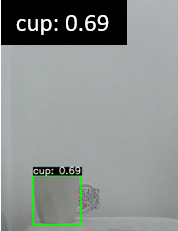} & 
\includegraphics[width=4.1cm, height=4.3cm]{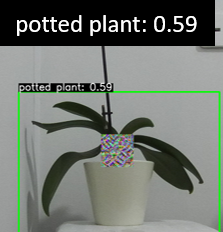} & \LARGE 0.77 \\ \cline{3-9} 
 &  & \LARGE Digital Learning - Physical Application & 
\includegraphics[width=4.1cm, height=4.3cm]{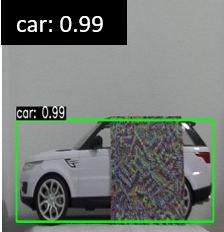} & 
\includegraphics[width=4.1cm, height=4.3cm]{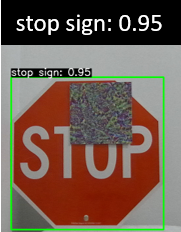} & 
\includegraphics[width=4.1cm, height=4.3cm]{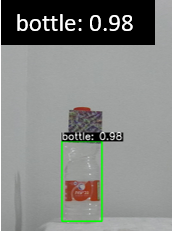} & 
\includegraphics[width=4.1cm, height=4.3cm]{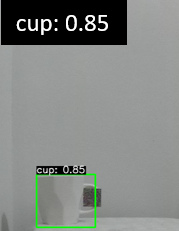} & 
\includegraphics[width=4.1cm, height=4.3cm]{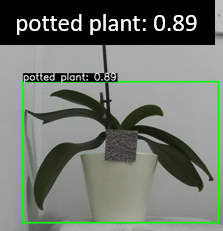} & \LARGE 0.93 \\ \hline
\end{tabular}%
}
\label{tab:physical_vs_digital}
\end{table*}


Here, we present the motivation to convert digital adversarial attacks for E2E execution in the physical domain.
We show that digitally learned patches may display reduced effectiveness when transferred to the physical domain as printed stickers. 
Specifically, in our experiments, we find that when patches are applied in the digital domain, the average confidence of object detectors decreases from 0.97 in the non-adversarial scenario to 0.66. 
However, when these patches are printed and applied in the physical domain, the average confidence only decreases to 0.90, demonstrating the challenge of maintaining an attack's effectiveness when transferring adversarial patches from the digital domain to the physical domain.
In Section \ref{subsec:challenges_in_transferability}, we review two key factors contributing to the difficulty of transferring adversarial patches from the digital to the physical domain. First, the printing process introduces color discrepancies, with an average of 99.34\% of the pixels differing between digital and printed patches. Second, environmental noise in the physical domain causes pixel values to vary significantly over short time intervals, even in controlled conditions.

\subsection{Experimental Setup}
\label{subsec:motivation_exp_setup}
We generated digital patches for five target objects: a car, a stop sign, a bottle, a cup, and a potted plant. These patches were generated using two untargeted adversarial attacks: DPatch~\cite{liu2018dpatch} and Robust DPatch~\cite{lee2019physical} (both from the ART~\cite{nicolae2018adversarial} library) against the YOLOv3 \cite{redmon2018yolov3} and Faster R-CNN \cite{ren2016faster} object detectors, respectively.
We selected Faster R-CNN and YOLOv3 because the DPatch and Robust DPatch attacks were originally designed and evaluated for them.

For each patch, we used the default settings defined in ART, specifically a learning rate of 5.0, batch size of 16, maximum iterations of 1000, and varying patch sizes depending on the size of the target object and attack implementation.
The patches were generated on a machine equipped with an NVIDIA RTX 2080 Ti GPU, six CPU cores, and 24 GB RAM.
We used the ZED2i camera to capture the objects from a distance of 0.5 meters. The same camera was used for both the patch learning and application phases.

For the physical domain application, we printed the patches as stickers on Chromo 300gsm paper using a Xerox Versant 280 printer (an industrial-grade printer typically found in professional print shops, with an approximate cost of \$35,000).
We evaluated the patches' effectiveness, \textit{i.e.,} the difference in the object detector's confidence when detecting the target object before and after applying the patch, in both the digital and physical domains.

\subsection{Failure of Transferability}
\label{sec:failure-of-transferability}
Table \ref{tab:physical_vs_digital} illustrates the effect of the digitally generated patches, both when applied digitally and when printed and placed in the physical domain, on the Faster R-CNN and YOLOv3 object detectors.
The average confidences of YOLOv3 and Faster R-CNN on the clean images without patches were 0.96 and 0.98, respectively.
When applying the DPatch and Robust DPatch attacks to YOLOv3 and Faster R-CNN digitally, the average confidences dropped to 0.55 and 0.77, respectively.
When printing and placing the patches in the physical domain, rather than applying the patches digitally, the average confidences of YOLOv3 and Faster R-CNN rose to 0.88 and 0.93, respectively. These results demonstrate that when patches generated in the digital domain are transferred to the physical domain, they may fail to perform as intended, since the object detectors continue to detect all objects with a high level of confidence.

\begin{insight} \label{insight:motivation1}
For the two object detectors evaluated, patches generated in the digital domain are not effective in the physical domain.
\end{insight}
\begin{insight} \label{insight:motivation2}
For the two adversarial attack methods evaluated, patches generated in the digital domain are not effective in the physical domain.
\end{insight}
\begin{insight} \label{insight:motivation3}
For the five targeted objects, patches generated in the digital domain are not effective in the physical domain.
\end{insight}


\begin{table}[]
\centering
\caption{The difference between patches applied digitally and the same patches applied physically as stickers.}
\begin{tabular}{|c|c|c|c|c|}
\hline
\textbf{Camera Model} & \textbf{Patch} & \boldmath{$L_{2}$} & \boldmath{$L_{\infty}$} & \boldmath{\textbf{$L_{0}$} (\%)} \\
\hline
\multirow{3}{*}{ZED2i} & \#1 & 15770.25 & 221 & 99.42 \\
                       & \#2 & 17249.71 & 213 & 99.50 \\
                       & \#3 & 13635.54 & 207 & 99.39 \\
\cline{1-5}
\multirow{3}{*}{iPhone 16} & \#1 & 11087.62 & 197 & 99.17 \\
                           & \#2 & 11329.53 & 204 & 99.06 \\
                           & \#3 & 12847.78 & 236 & 99.30 \\
\cline{1-5}
\multirow{3}{*}{YI Dash Camera} & \#1 & 15779.49 & 231 & 99.39 \\
                           & \#2 & 16285.12 & 239 & 99.40 \\
                           & \#3 & 13981.11 & 242 & 99.46 \\
\cline{1-5}
\multicolumn{2}{|c|}{\textbf{Average}} & 14218.46 & 221.11 & 99.34 \\
\hline
\end{tabular}
\label{tab:motivation_patch_comparison}
\vspace{-1.5em}
\end{table}

\subsection{Cause of the Failure}
\label{subsec:challenges_in_transferability}
Here, we conduct two experiments to explore the causes of the transferability issue between digital and physical domains. 

In Section \ref{subsubsec:digital_vs_physical_as_stickers} we compare digital patches with their printed counterparts as stickers, analyzing how they are perceived by the camera lens. 

In Section \ref{subsubsec:environmental_noise} we evaluate the impact of environmental noise on recorded consecutive frames. We captured multiple images of the same object at short intervals of 30 seconds in a controlled environment to examine the variability in pixel values over time.

Both experiments utilized three cameras from different categories: a smartphone camera (iPhone 16), a stereo camera (ZED2i), and a dash camera (YI Smart Dash Camera with ADAS capabilities).
For the first experiment, we used three different patches generated during the experiment described in Section \ref{subsec:motivation_exp_setup}.

\subsubsection{Differences Between Digital and Physical Patches}
\label{subsubsec:digital_vs_physical_as_stickers}
One key factor in the transferability issue is the difficulty in accurately reproducing the colors of digital patches when printing them for physical application. This is evident in Table \ref{tab:physical_vs_digital}, which shows that identical digital patches appear different when applied digitally versus physically, even when using a professional industrial printer (Xerox Versant 280, priced at approximately \$35,000).

Table \ref{tab:motivation_patch_comparison} shows the pixel-level differences for three different digital patches and their printed counterparts. On average, 99.34\% of the pixels differed between respective digital and printed patches, highlighting the inconsistencies caused by the printing process.

\begin{insight} \label{insight:motivation4}
The difference between digital and printed colors influences patch transferability. Patches that are identical in the digital domain, when printed, demonstrate significant pixel-level discrepancies compared to their digital versions.
\end{insight}

\subsubsection{Differences Between Consecutive Captures of the Same Scene}
\label{subsubsec:environmental_noise}

\begin{table}[]
\centering
\caption{Comparison of pixel differences between consecutive images captured at 30-second intervals under consistent and controlled conditions.}
\resizebox{\columnwidth}{!}{%
\begin{tabular}{|l|l|l|l|l|l|}
\hline
\textbf{\makecell{Camera \\ Model}} & \textbf{\makecell{Previous \\ Image}} & \textbf{\makecell{Current \\ Image}} & \boldmath{\centering $L_{2}$} & \boldmath{\centering $L_{\infty}$} & \boldmath{\centering $L_{0}$ (\%)} \\ \hline
\multirow{4}{*}{ZED2i} & 
\makebox[0.18\linewidth]{\raisebox{-0.5\height}{\includegraphics[width=0.18\linewidth]{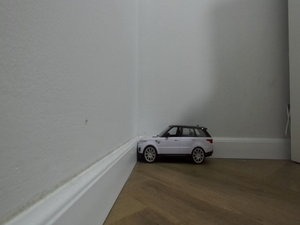}}} & 
\makebox[0.18\linewidth]{\raisebox{-0.5\height}{\includegraphics[width=0.18\linewidth]{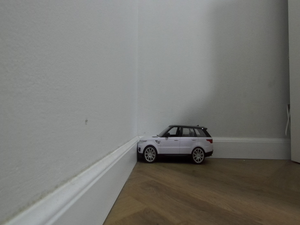}}} & 
10371.30 & 45 & 89.69 \\ \cline{2-6}
                       & 
\makebox[0.18\linewidth]{\raisebox{-0.5\height}{\includegraphics[width=0.18\linewidth]{figs/limitations/Zed2i/rsz_20241228_184415_left.png}}} & 
\makebox[0.18\linewidth]{\raisebox{-0.5\height}{\includegraphics[width=0.18\linewidth]{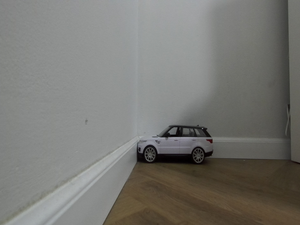}}} & 
9981.69 & 85 & 89.59 \\ \cline{2-6}
                       & 
\makebox[0.18\linewidth]{\raisebox{-0.5\height}{\includegraphics[width=0.18\linewidth]{figs/limitations/Zed2i/rsz_20241228_184446_left.png}}} & 
\makebox[0.18\linewidth]{\raisebox{-0.5\height}{\includegraphics[width=0.18\linewidth]{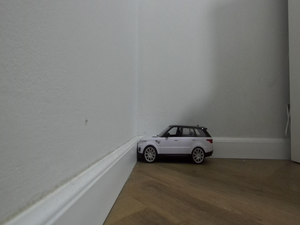}}} & 
9844.01 & 87 & 89.22 \\ \hline
\multirow{4}{*}{iPhone 16} & 
\makebox[0.18\linewidth]{\raisebox{-0.5\height}{\includegraphics[width=0.18\linewidth]{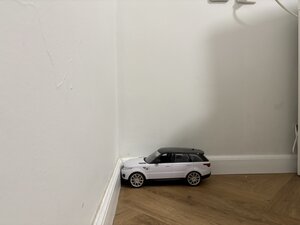}}} & 
\makebox[0.18\linewidth]{\raisebox{-0.5\height}{\includegraphics[width=0.18\linewidth]{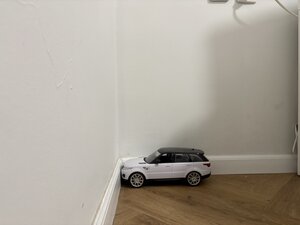}}} & 
4595.37 & 33 & 79.23 \\ \cline{2-6}
                       & 
\makebox[0.18\linewidth]{\raisebox{-0.5\height}{\includegraphics[width=0.18\linewidth]{figs/limitations/iphone16/rsz_11img_0806.jpg}}} & 
\makebox[0.18\linewidth]{\raisebox{-0.5\height}{\includegraphics[width=0.18\linewidth]{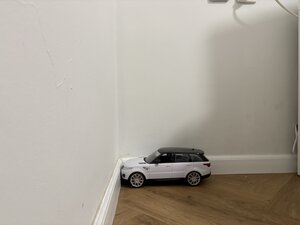}}} & 
4075.02 & 35 & 68.28 \\ \cline{2-6}
                       & 
\makebox[0.18\linewidth]{\raisebox{-0.5\height}{\includegraphics[width=0.18\linewidth]{figs/limitations/iphone16/rsz_1img_0807.jpg}}} & 
\makebox[0.18\linewidth]{\raisebox{-0.5\height}{\includegraphics[width=0.18\linewidth]{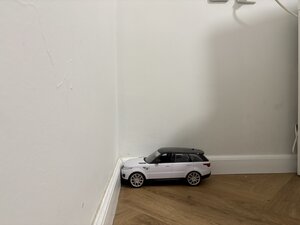}}} & 
5970.31 & 78 & 73.33 \\ \hline
\multirow{4}{*}{\makecell{YI \\ Dash \\ Camera}} & 
\makebox[0.18\linewidth]{\raisebox{-0.5\height}{\includegraphics[width=0.18\linewidth]{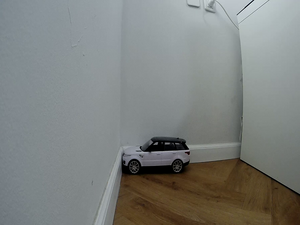}}} & 
\makebox[0.18\linewidth]{\raisebox{-0.5\height}{\includegraphics[width=0.18\linewidth]{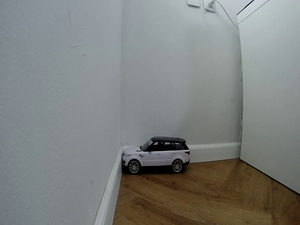}}} & 
10340.12 & 125 & 71.85 \\ \cline{2-6}
                       & 
\makebox[0.18\linewidth]{\raisebox{-0.5\height}{\includegraphics[width=0.18\linewidth]{figs/limitations/YI/rsz_1frame_00-30.png}}} & 
\makebox[0.18\linewidth]{\raisebox{-0.5\height}{\includegraphics[width=0.18\linewidth]{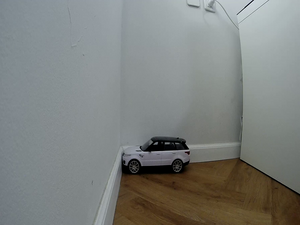}}} & 
9762.41 & 96 & 71.78 \\ \cline{2-6}
                       & 
\makebox[0.18\linewidth]{\raisebox{-0.5\height}{\includegraphics[width=0.18\linewidth]{figs/limitations/YI/rsz_frame_01-00.png}}} & 
\makebox[0.18\linewidth]{\raisebox{-0.5\height}{\includegraphics[width=0.18\linewidth]{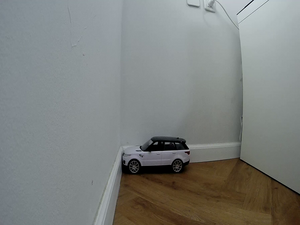}}} & 
9439.34 & 88 & 71.35 \\ \hline
\multicolumn{3}{|c|}{\textbf{Average}} & 8264.40 & 74.67 & 78.26 \\ \hline
\end{tabular}%
}
\label{tab:image_diff}
\vspace{-1.5em}
\end{table}

An additional factor affecting patch transferability is the noise introduced from recording in the physical domain. While in the digital domain the environment is stable (only the patch changes over time), in the physical domain the majority of pixels in the image change over time, adding noise that can affect patches' performance.
This is demonstrated in Table~\ref{tab:image_diff}. For three camera models, four images of the same scene are captured in a controlled laboratory environment under constant conditions at 30-second intervals.
No projections (e.g., performing PAPLA) were made during the recording.
For each camera, we measured the $L_{2}$, $L_{\infty}$, and $L_{0}$ values of consecutive images.
We found high values of $L_{2}$, $L_{\infty}$, and $L_{0}$ between consecutive images, averaging 8264.40, 74.67, and 78.26\%, respectively, which demonstrate that external environmental factors introduce significant noise in the physical domain, even under controlled conditions.

\begin{insight} \label{insight:motivation5}
The difference between pixel values in the digital and physical domains can influence patch transferability.
A patch optimized for a specific digital scene may not perform as expected when deployed in a physical scene (even if identical) due to external environmental factors affecting the recorded pixel values.
\end{insight}

These findings and insights indicate that in order to ensure the success of adversarial attacks, the learning process for adversarial patches must be performed in the physical domain.
This is because adversarial learning in the digital domain does not ensure similar performance when applied in the physical domain.
\section{Threat Model and Method}
\label{sec: threat_model-method}







Here, we outline the threat model and methodology for conducting end-to-end (E2E) adversarial learning in the physical domain. 
We define the adversary’s capabilities, justify the approach, and detail PAPLA, our framework for generating and applying adversarial patches under real-world conditions in the physical domain.

\begin{figure} 
	\centering
	\includegraphics[width=0.98\linewidth]{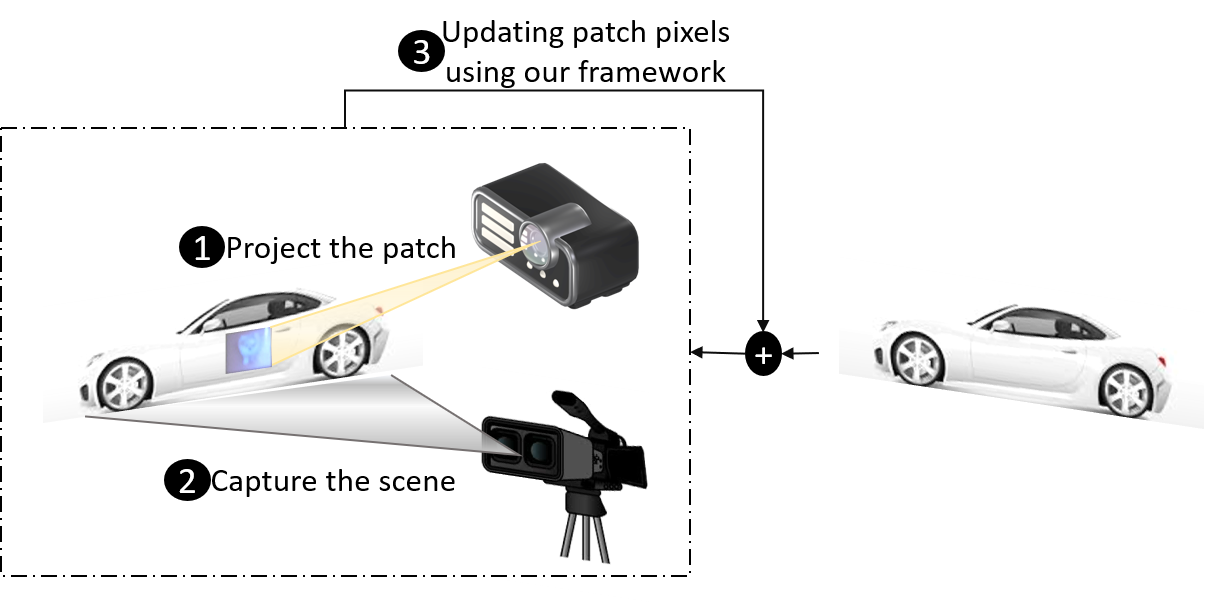}
	\caption{PAPLA learning process: an adversary points a projector and a camera at the target object and (1) projects a patch onto the object, as well as (2) captures the scene that contains the object with the projected patch. (3) The patch pixels are updated using PAPLA, and the process repeats.}
	\label{fig:threat_model}
\end{figure}

\subsection{Threat Model}
We assume that an adversary is interested in performing an evasion attack in the physical domain to hide an object from an object detector.
The purpose of the attack is to produce a patch through physical domain learning, and thus ensure the hiding of an object from the object detector in the physical environment.
Furthermore, we assume that the target object is static and has a suitable surface for projection. A potential use case for this scenario is hiding parked vehicles and road signs from the detection systems of autonomous vehicles.
 
\subsubsection{Attacker's Capabilities and Knowledge}
We assume that the adversary has access to a position with a visual line of sight to the target object (the object that the attacker wants to hide) to allow projection of the adversarial patch. The target object is assumed to possess a suitable surface for projecting a patch. In addition, we assume that they can place a projector to project a patch on the target object, as well as a camera to capture the scene with the projected patch (see Figure \ref{fig:threat_model}).

\subsubsection{Extension of Previously Evaluated Threat Model}
We note that state-of-the-art methods have adopted a similar threat model, employing projectors to apply adversarial attacks in the physical domain \cite{lovisotto2021slap,nassi2020phantom,wen2024opticloak,hu2023adversarial}. Our approach extends this threat model by assuming that the adversarial learning process itself can also be conducted entirely in the physical domain using a projector.

\subsubsection{Significance}
The significance of our threat model is that, unlike previous methods where adversarial learning and attack application were conducted in different domains (e.g., \cite{athalye2018synthesizingrobustadversarialexamples,song2018physical,lee2019physical,eykholt2018robust,chen2019shapeshifter,liu2018dpatch,zhang2018camou,thys2019fooling,xu2020adversarial,huang2020universal,wu2020making,zolfi2021translucent,hu2021naturalistic,jing2021too,tan2021legitimate,suryanto2022dta,hu2022adversarial,jia2022fooling,huang2023t,zhu2023tpatch,hu2023physically,guesmi2024dap,weirevisiting, chengfull,zhu2022infrared,wei2023hotcold,wei2024infrared,zhu2024infrared,lovisotto2021slap,wen2024opticloak,hu2023adversarial,hwang2023adversarial,komkov2021advhat,lin2022real}, we conduct both adversarial learning and attack application in the physical domain.
This ensures that the challenges associated with transferability between the digital and physical domains do not degrade the performance of the attack, thereby ensuring the success of the learning process to imply the success of the attack.

\subsection{Method - PAPLA}
Here we review the methodology of the Physical-domain Adversarial Patch Learning Augmentation (PAPLA) framework, employed to conduct E2E adversarial patch learning in the physical domain.
Our approach is to adapt existing adversarial methods, where the learning process is carried out in the digital domain, to be carried out E2E in the physical domain.
Unlike prior methods, where the patch learning phase occurs before the application phase, our approach performs both phases simultaneously.
E2E adversarial patch learning in the physical domain deploys the latest iteration's patch at the target object and, according to the utilized adversarial method, updates the patch given the physical domain conditions and redeploys it.
This improves the performance of the patch application phase since it is integrated in the learning process.
PAPLA's adversarial patch learning process, as illustrated in Figure~\ref{fig:threat_model}, is as follows.

\begin{enumerate}
    \item \textbf{Iterative Learning:} In this phase, the framework generates a random digital patch using an existing digital attack method (e.g., DPatch~\cite{liu2018dpatch} or NAP~\cite{hu2021naturalistic}). The patch is then iteratively optimized according to the chosen attack, using footage from the physical domain:
    \begin{enumerate}
        \item Apply (project) the patch onto the target object using a projector.
        \item Capture the physical scene containing the object and the projected patch with a camera.
        \item Update the patch pixels iteratively using the chosen attack, leveraging the physical conditions to maximize the adversarial effect.
    \end{enumerate}
    \item \textbf{Attack Application:} After the patch is fully optimized, it is projected onto the target object in the physical environment to mislead the object detector.
\end{enumerate}

PAPLA wraps existing adversarial attack methods, originally designed for adversarial learning in the digital domain, and transforms them into a framework for physical-domain adversarial learning. This transformation is achieved by projecting the patches onto the target object and iteratively capturing images of the physical scene using a camera. By integrating the physical conditions directly into the learning process, PAPLA ensures that the adversarial attack is optimized for real-world scenarios.
\section{Analysis}
\label{sec:analysis}
Here, we analyze the factors influencing the success of end-to-end (E2E) adversarial learning in the physical domain. We investigate the impact of environmental conditions and surface characteristics. We aim to identify the key elements that contribute to the effectiveness and robustness of adversarial patches under various physical-domain learning scenarios, as well as physical constraints under which E2E adversarial learning in the physical domain demonstrates reduced effectiveness.
The experiments presented in this section were conducted in a controlled laboratory environment to allow us to control and isolate the examined factors. 

\begin{figure*}
	\centering
	\includegraphics[width=0.85\linewidth]{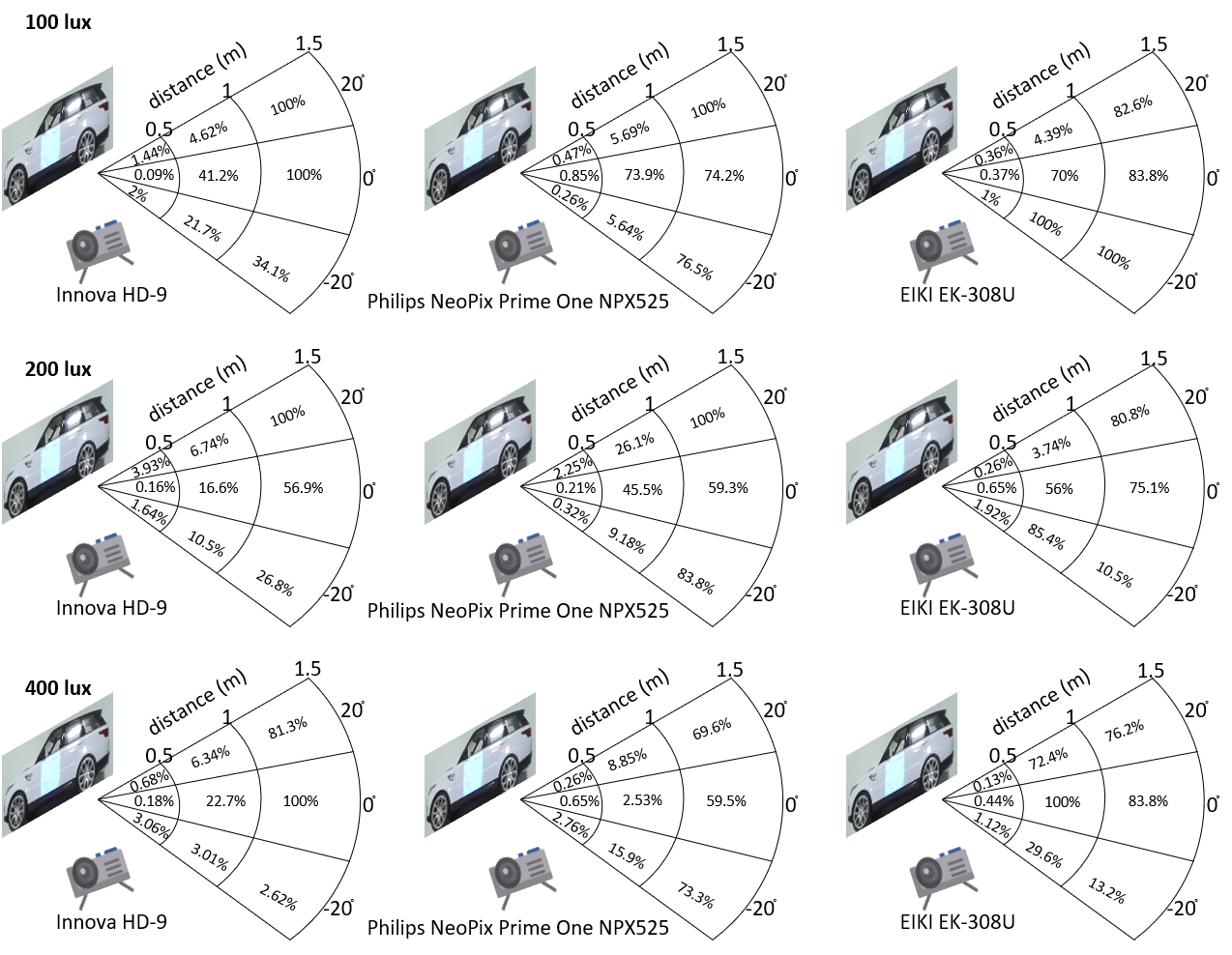}
	\caption{Confidence reduction percentage for different angles, distances, ambient light levels, and projectors. Each cell shows the percentage difference between the original confidence score (without patch projection) and the confidence score with patch projection learned E2E in the physical domain.}
	\label{fig:analysis_1}
\end{figure*}

\subsection{Impact of Environmental Factors on Attack Success}

Here, we analyze how various environmental factors affect PAPLA's success in improving confidence score reduction. 
Specifically, we perform the Dpatch \cite{liu2018dpatch} attack on a car against YOLOv3, with the learning process carried out in the physical domain, and examine the effect of the following factors: projector strength, ambient light (measured in lux), distance, and the angle of the target object in relation to the camera.
We test the effect of each factor in combination with the other factors. 
In total, we performed 81 different runs to analyze each factor's impact on the attack's success.

\subsubsection{Experimental Setup}
We examine several values for each environmental factor (projector strength, ambient light, distance, and target object angle) to analyze the impact of each factor on PAPLA's success.
Figure~\ref{fig:analysis_1} provides an overview of the experimental setup.
We used three types of projectors: Innova HD-9 with a light output of 1800 ANSI lumens, Philips NeoPix Prime One NPX525 with a light output of 3000 ANSI lumens, and EIKI EK-308U with a light output of 6000 ANSI lumens.
Additionally, we tested three ambient light levels: 100 lux, 200 lux, and 400 lux, measured using the Extech HD450 light meter. Furthermore, we analyzed three distances and three different angles of the camera in relation to the object. 
For distances, we analyzed 0.5 meters, 1 meter, and 1.5 meters. 
For angles, we analyzed $0\degree$, $20\degree$, and $-20\degree$ ($340\degree$).
We used the default parameters specified for the DPatch attack in the ART library.
The camera used was the ZED 2i Stereo Camera.
The size of the car was \textit{33×11cm}, and the size of the patches in this analysis was \textit{4.5×4.5cm}.

\begin{figure*}[]
\centering
\setlength{\abovecaptionskip}{4pt} 
\setlength{\belowcaptionskip}{0pt} 
\begin{subfigure}[b]{0.24\textwidth}
    \includegraphics[width=\linewidth]{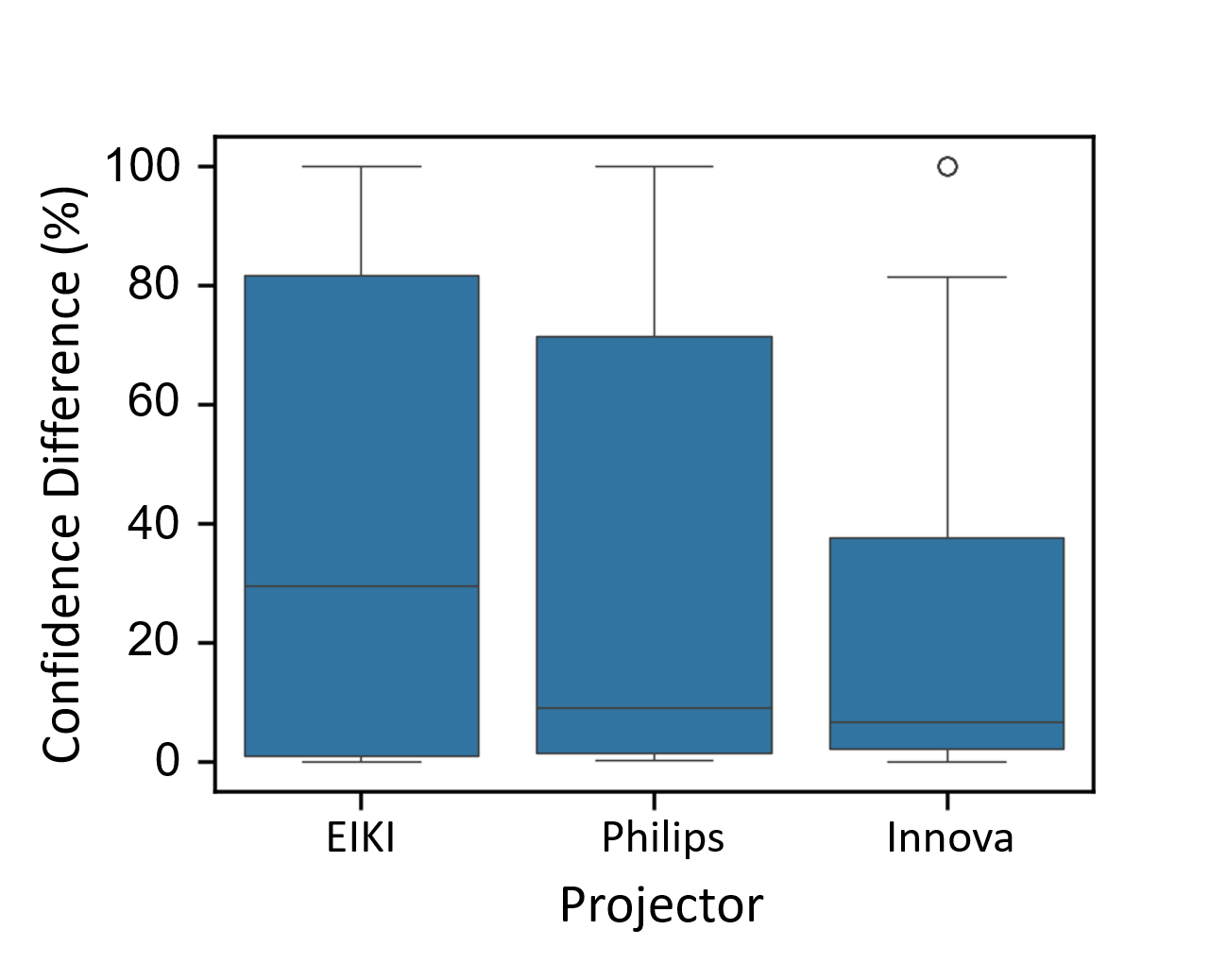}
    \caption{Impact of Projector.}
    \label{fig:projector_impact}
\end{subfigure}
\begin{subfigure}[b]{0.24\textwidth}
    \includegraphics[width=\linewidth]{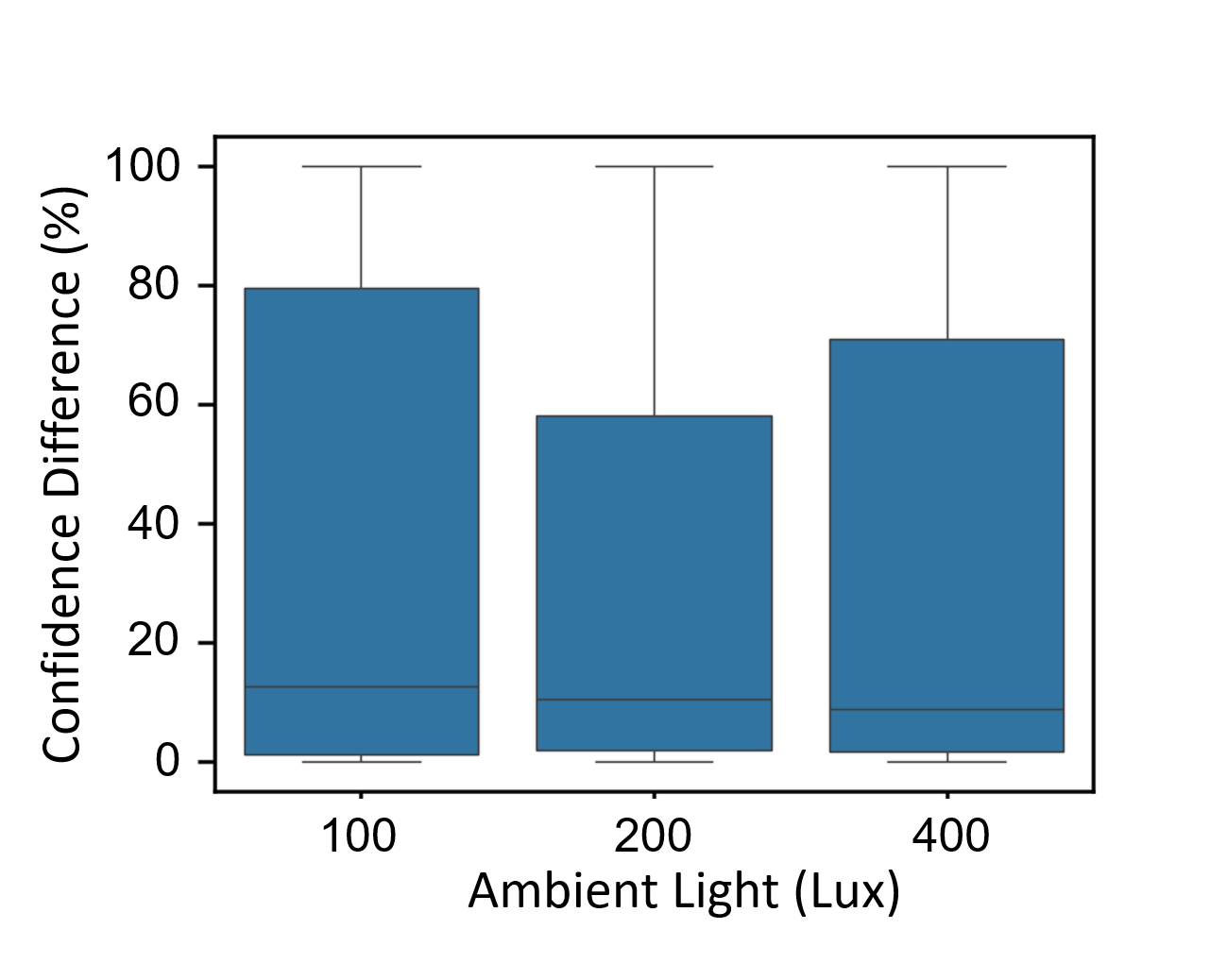}
    \caption{Impact of Ambient Light.}
    \label{fig:light_impact}
\end{subfigure}
\begin{subfigure}[b]{0.24\textwidth}
    \includegraphics[width=\linewidth]{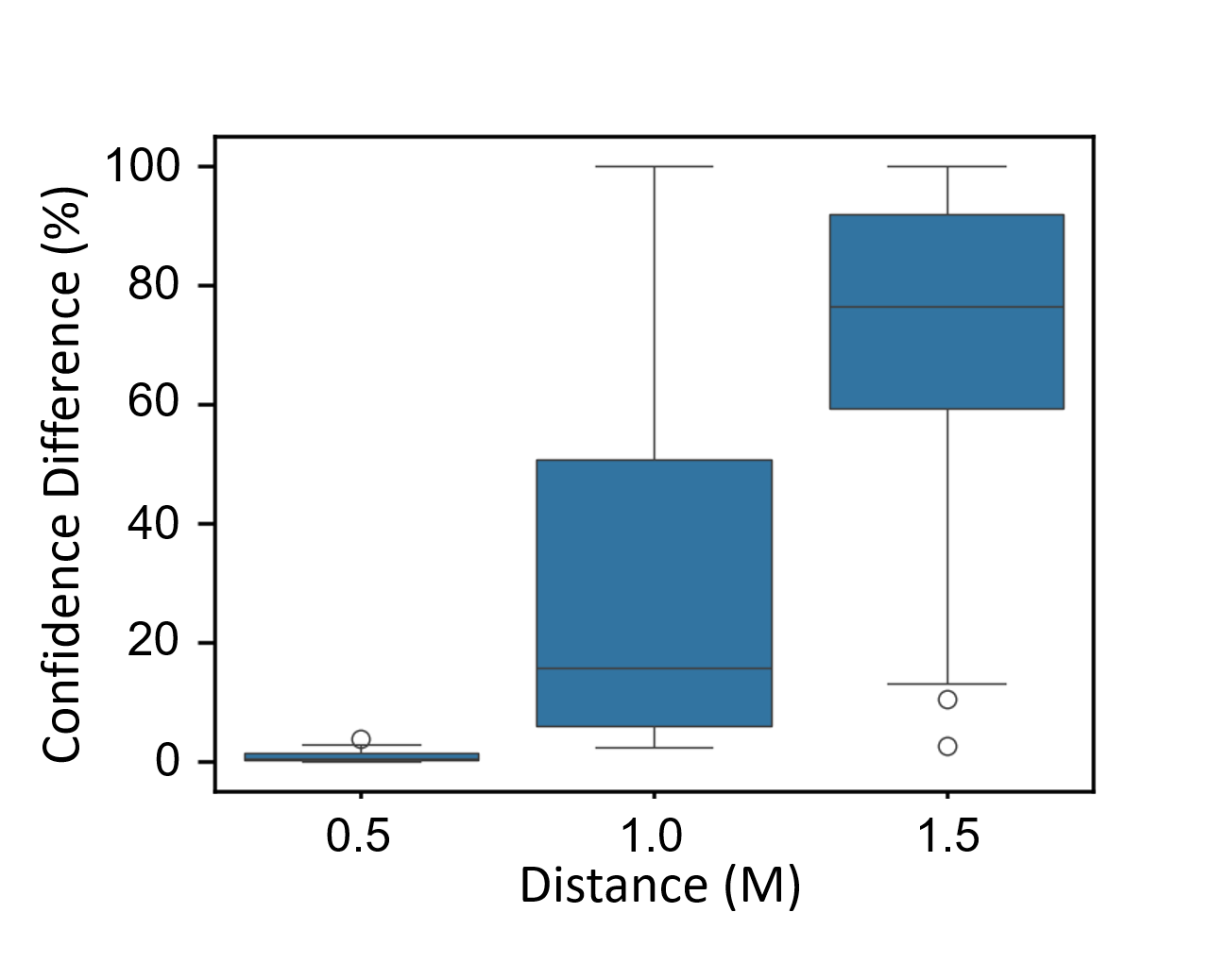}
    \caption{Impact of Distance.}
    \label{fig:dis_impact}
\end{subfigure}
\begin{subfigure}[b]{0.24\textwidth}
    \includegraphics[width=\linewidth]{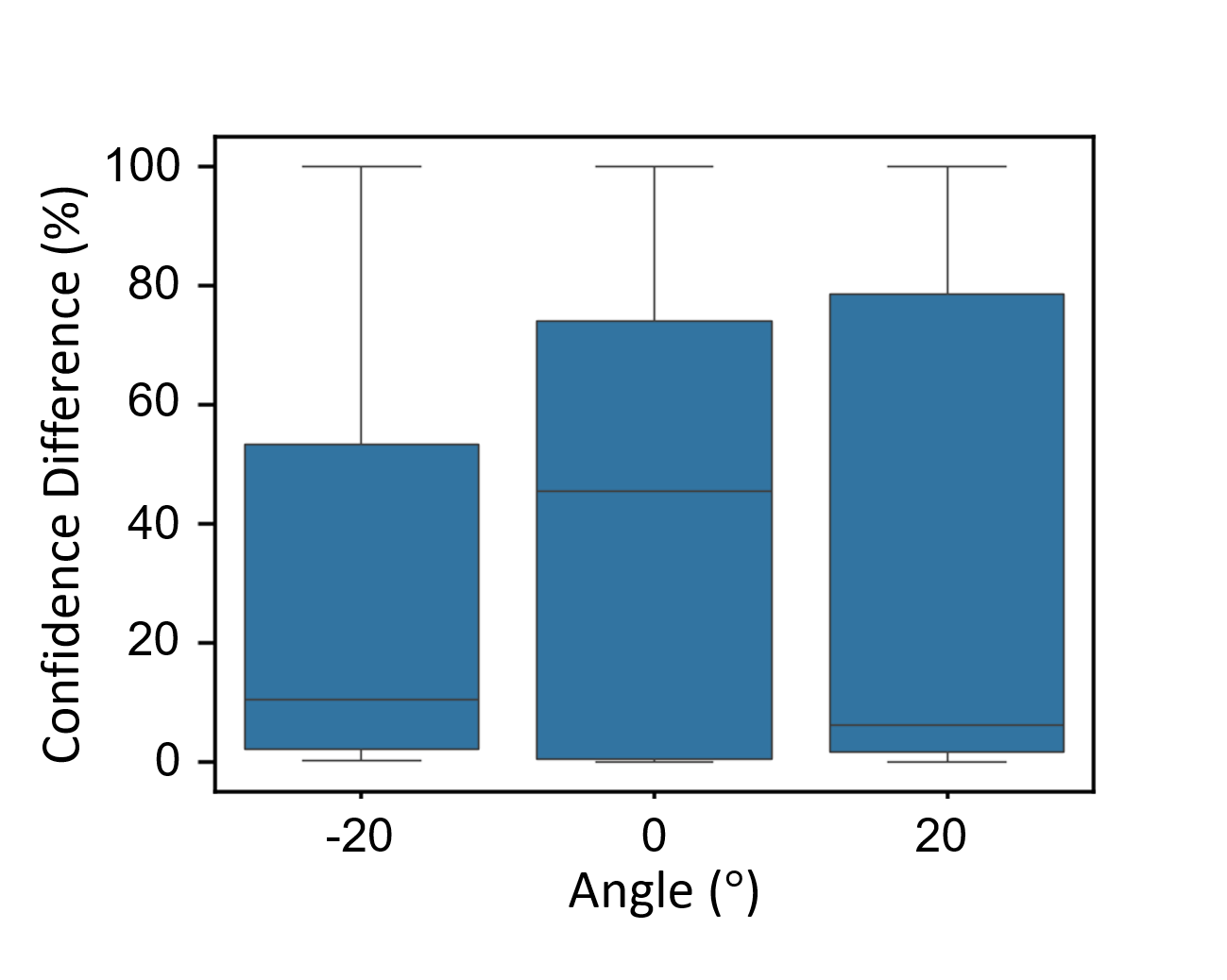}
    \caption{Impact of Angle.}
    \label{fig:ang_impact}
\end{subfigure}

\caption{Box plots illustrating the impact of each environmental factor on the confidence reduction percentage of the DPatch attack performed using PAPLA (E2E in the physical domain). The Y-axis represents the percentage difference between the original confidence score (without patch projection) and the confidence score with patch projection using PAPLA.}
\label{fig:analysis_box_plots}
\end{figure*}

\subsubsection{Results}
The results of our experiments, presented in Figure~\ref{fig:analysis_1}, highlight several key insights into the factors affecting the confidence reduction percentage of PAPLA on a car in the physical domain. To understand the effect of each factor, we used box plots (see Figure~\ref{fig:analysis_box_plots}) and ANOVA analysis, which provided clear visualization and statistical significance of the impact of projector strength, distance, angle, and ambient light on the confidence reduction percentage.

\textbf{Projector Strength.}
The projector strength appears to have a moderate impact on PAPLA's confidence reduction percentage. The ANOVA results show a p-value of approximately 0.0908, indicating a trend approaching significance, but not a strong effect. Figure \ref{fig:projector_impact} shows that the EIKI projector has the highest median confidence reduction percentage at approximately 29.60\%, followed by the Philips projector at 9.18\%, and the Innova projector at 6.74\%. This suggests that the EIKI projector, with the highest light output of 6000 ANSI lumens, is the most effective.

\vspace{2mm} 
\begin{insight} \label{insight:proj_impact}
Higher projector light output improves the performance (confidence reduction percentage) of PAPLA.
\end{insight}

\textbf{Ambient Light.}
Ambient light appears to have a relatively minor impact on the confidence reduction percentage. The ANOVA results show a p-value of approximately 0.332, indicating a non-significant effect. Figure~\ref{fig:light_impact} indicates overlapping distributions for different lux levels, with medians of 12.67\% for 100 lux, 10.56\% for 200 lux, and 8.86\% for 400 lux. This indicates no clear trend of ambient light levels significantly affecting the confidence reduction percentage.

\vspace{2mm} 
\begin{insight} \label{insight:light_impact}
PAPLA is robust to varying lighting conditions, making the patches effective across different ambient light levels.
\end{insight}

\textbf{Camera Distance.}
The distance of the camera from the attacked object has a significant impact on the confidence reduction percentage. The ANOVA results show a very low p-value (approximately $2.35e^{-15}$), indicating a highly significant effect. Figure~\ref{fig:dis_impact} shows that the median confidence reduction percentage increases substantially with distance: 0.65\% at 0.5 meters, 15.91\% at 1 meter, and 76.53\% at 1.5 meters.
We note that beyond the tested range (greater than 1.5 meters), the object is not detected at all, regardless of whether a patch is projected or not.

\vspace{2mm} 
\begin{insight} \label{insight:dis_impact}
Within the tested distance range (0.5m to 1.5m), increasing the distance between the camera and the target object improves PAPLA's performance (confidence reduction percentage).
\end{insight}

\textbf{Angle.}
The angle of the camera in relation to the attacked object shows a moderate effect on the PAPLA's confidence reduction percentage. The ANOVA results provide a p-value of approximately 0.112, indicating some effect but not highly significant. Figure~\ref{fig:ang_impact} shows that the median confidence reduction percentage is highest at $0\degree$ (45.49\%), followed by $-20\degree$ (10.51\%) and $20\degree$ (6.34\%). This suggests that a $0\degree$ angle tends to yield a higher confidence reduction percentage.

\vspace{2mm} 
\begin{insight} \label{insight:ang_impact}
The camera angle has a moderate impact on PAPLA's confidence reduction percentage, with the optimal performance generally seen at $0\degree$.
\end{insight}

\subsection{Effect of Surface Color on Patch Effectiveness}
\label{subsec:surface_analysis}
In this part, we analyze the impact of surface color on PAPLA's effectiveness in reducing object detection confidence.


\begin{figure}[]
	\centering
	\includegraphics[width=0.98\linewidth]{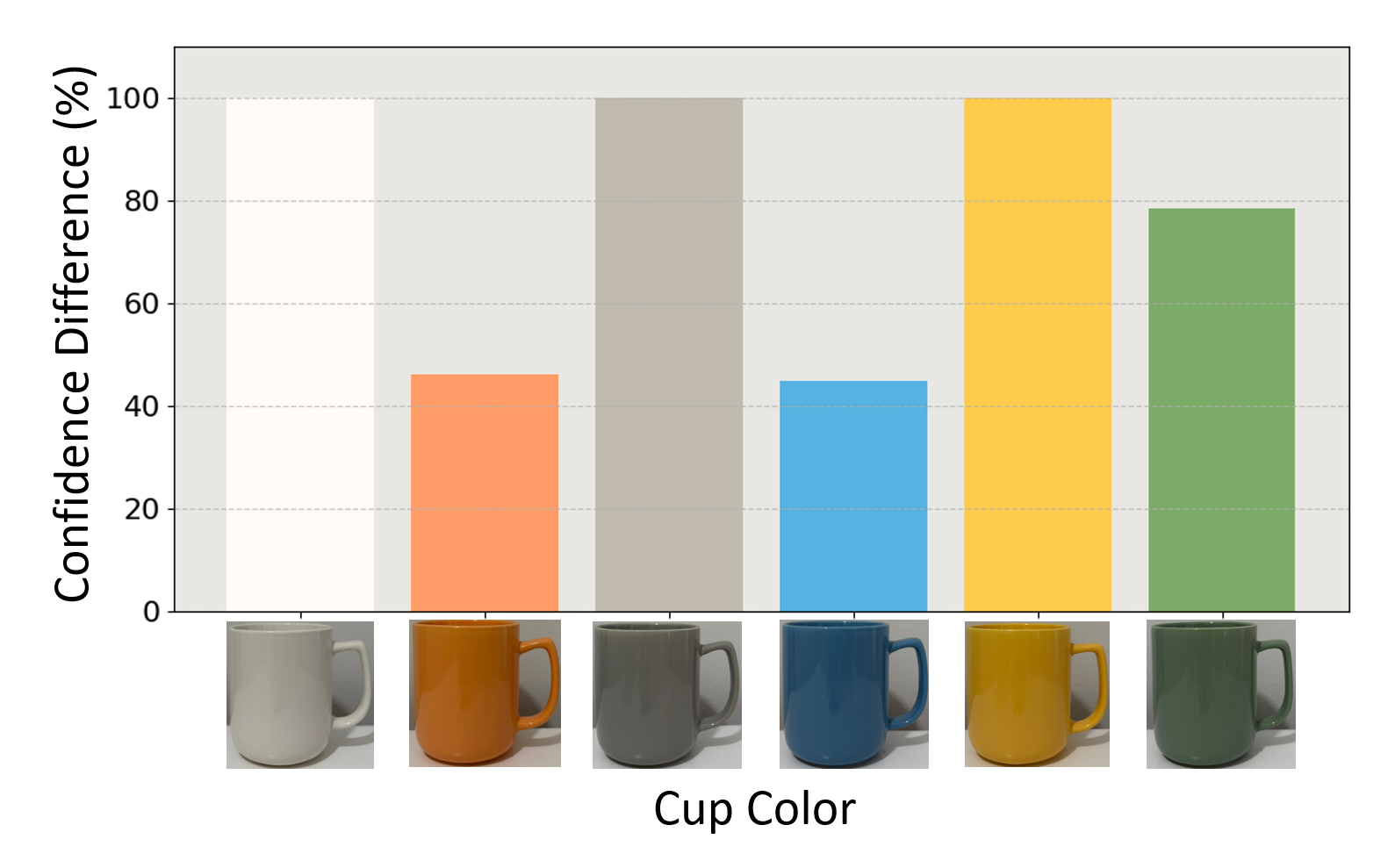}
	\caption{Impact of surface color on patch projection effectiveness: Each bar corresponds to a cup of a specific color, indicated by the bar's color. The Y-axis shows the percentage decrease in the object detection model's confidence score when a patch is projected onto the cup, compared to the confidence score without the patch. 
    }
	\label{fig:surface_analysis}
\end{figure}

\subsubsection{Experimental Setup}
To demonstrate the projection surface color's effect on PAPLA's effectiveness, we conducted the NAP~\cite{hu2021naturalistic} attack using PAPLA's E2E physical domain learning framework. The attack was applied to six identical ceramic cups, with the only difference being their color, as seen in Figure~\ref{fig:surface_analysis}. For each color, we measured the percentage difference in the confidence score returned by the Faster R-CNN object detector between the clean cup (without patch projection) and the cup with the adversarial patch projected and learned E2E in the physical domain using PAPLA. For each color, we ran the attack three times and averaged the results.

This analysis was performed on a TITAN X Pascal machine with four CPU cores and 32 GB of RAM. A Stereolabs ZED2i camera was used to capture the scene, while an EIKI EK-308U projector was used to project the patch in the physical domain. The same parameters were applied across all scenarios, following the default settings from the original attack implementation, with a slight modification: each patch learning process was limited to 50 epochs instead of the original 100. Environmental conditions, including camera position, distance, patch size, and ambient lighting, were kept constant throughout all runs. The cup was placed 0.85 meters from the camera at a $0\degree$ angle, with a \textit{4×4cm} patch applied. The ambient light was maintained at 100 lux, measured using an Extech HD450 light meter.

\subsubsection{Results}
The results, as illustrated in Figure~\ref{fig:surface_analysis}, demonstrate the impact of the color of the projection surface on the success of the adversarial patch attack.
The lighter surfaces (white, light grey, and yellow) yielded the most significant reductions, achieving a 100\% confidence score drop. In contrast, darker surfaces (green, orange, and blue) showed reduced effectiveness, with confidence score decreases of 78.44\%, 46.22\%, and 44.76\%, respectively. This suggests that lighter colors allow the adversarial patch to be projected more clearly, enhancing its effectiveness.
While PAPLA reduces detection confidence across all tested surfaces, the results highlight a variation in performance according to the color of the surface.

\vspace{2mm} 
\begin{insight} \label{insight:surface_impact} PAPLA's success at reducing detection confidence is highly effected by the color of the surface, with lighter surfaces yielding better results. \end{insight}





\section{Evaluations}
\label{sec:evaluations}

In this section, we review the evaluations conducted for our proposed framework PAPLA. 
All evaluations (except Section~\ref{subsec:outside_env}) were performed in a controlled laboratory environment to maintain consistency and reproducibility. The final evaluation, presented in Section~\ref{subsec:outside_env}, was conducted in an outdoor environment to evaluate the framework's performance in realistic conditions.
Specifically, we evaluate the DPatch \cite{liu2018dpatch} and NAP \cite{hu2021naturalistic} attacks after applying PAPLA. 
We evaluate the effectiveness of these attacks in three scenarios. In the \textbf{digital learning-digital application (DL-DA)} scenario, both patch learning and attack execution occur in the digital domain. In the \textbf{digital learning-physical application (DL-PA)} scenario, the patch is learned in the digital domain and applied as a printed sticker in the physical domain. Finally, in the \textbf{physical learning-physical application (PL-PA)} scenario, which is our proposed framework, both the patch learning and attack execution are performed E2E in the physical domain. We use the DL-PA scenario as a baseline to compare with PAPLA's physical domain performance. We also evaluate patch transferability across different object detectors.

\subsection{Robustness Against Various Target Objects}
\label{subsec:various_objs}

Here, we evaluate the robustness of PAPLA's ability to improve the performance of DL-PA adversarial attack methods on various \textit{target objects}.

\subsubsection{Experimental Setup}
\label{subsubsec:various_objs_exp_setup}
We evaluate PAPLA on four target objects: a stop sign, a car, a potted plant, and a cup.
All evaluations were performed on a TITAN X Pascal machine with four CPU cores and 32 GB RAM. We used a camera with two lenses for the physical domain experiments: the Stereolabs ZED2i to capture the scene and an EIKI EK-308U projector to project the patch in the physical domain. The digital patches were printed as stickers on Chromo 300gsm paper using a Xerox Versant 280 printer.
We used identical parameters in each scenario, specifically the default settings defined in the original attack implementations. We also ensured consistency in the environmental conditions, including ambient lighting which was maintained at 100 lux and measured using an Extech HD450 light meter. Camera position, distance, and patch size were consistent across both digital and physical domains. The car and stop sign were captured from a distance of 1.5 meters, the potted plant was captured from 0.6 meters, and the cup was captured from 0.5 meters; these distances were chosen to ensure a high confidence score from the object detector (around 1.0) for the clean objects. All objects were captured from a $0\degree$ angle relative to the camera. The sizes of the objects are as follows: the car was \textit{33×11cm}, the stop sign was \textit{30×30cm}, the potted plant was \textit{34x20cm}, and the cup was \textit{10×11cm}. The patch sizes were \textit{4.5×4.5cm} for the car, \textit{10×10cm} for the stop sign, \textit{4x4cm} for the potted plant, and \textit{4×4cm} for the cup. The recorded scene in the physical domain was identical to that in the digital domain.
PAPLA was evaluated in both stereoscopic and monocular scenarios. In the monocular scenario, we evaluate the confidence score returned by the object detector for each target object observed by a single lens. In the stereoscopic scenario, we use the highest confidence score returned from the two lenses of the camera, allowing us to determine whether the object is evaded in a scene that involves communication between the stereoscopic camera and the object detector.

\begin{figure*}[]
\centering
\begin{subfigure}[b]{0.49\textwidth}
    \includegraphics[width=\linewidth]{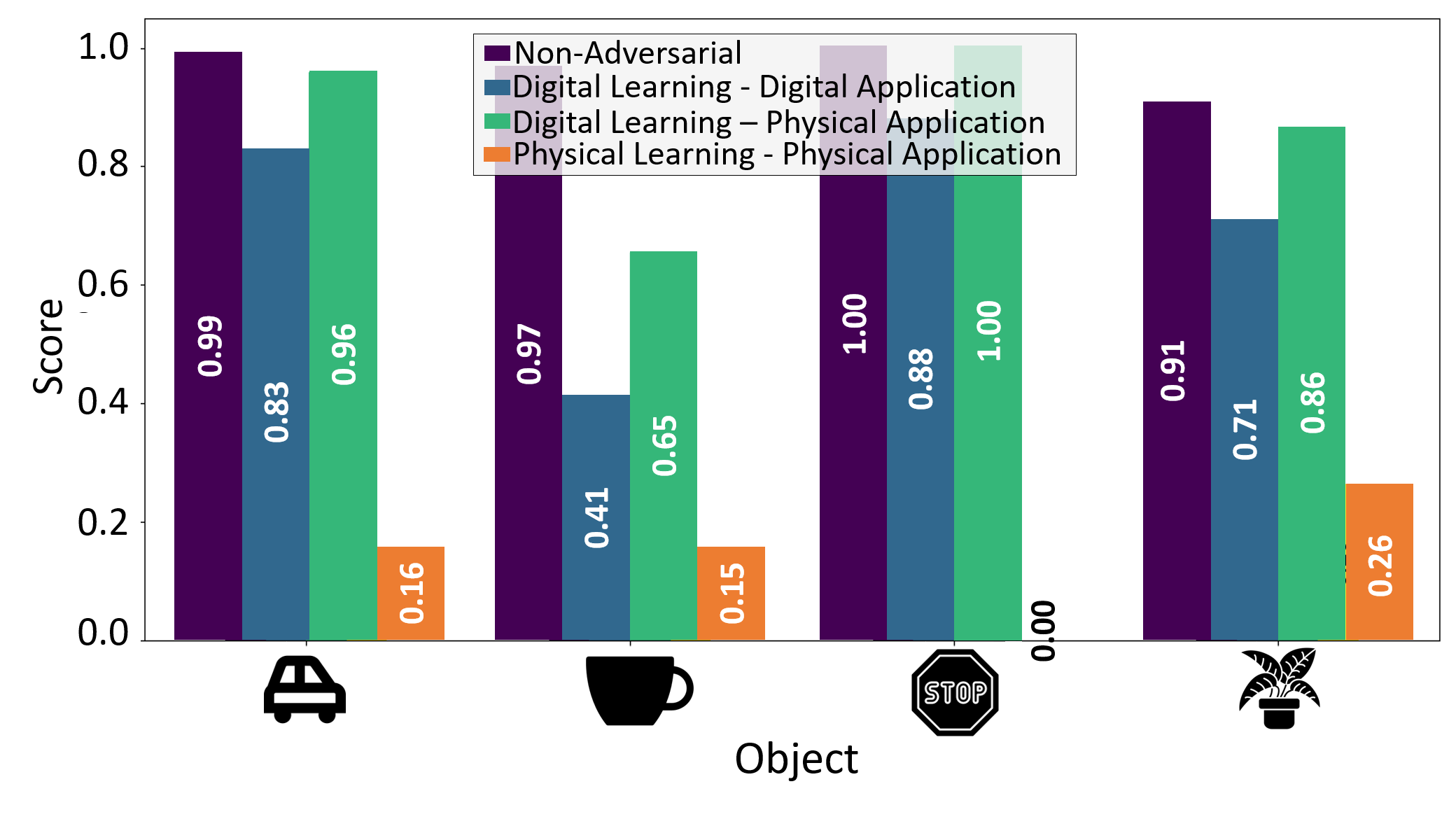}
    \caption{DPatch against YOLOv3 using a monocular camera.}
    \label{fig:dpatch_monocular}
\end{subfigure}
\hfill
\begin{subfigure}[b]{0.49\textwidth}
    \includegraphics[width=\linewidth]{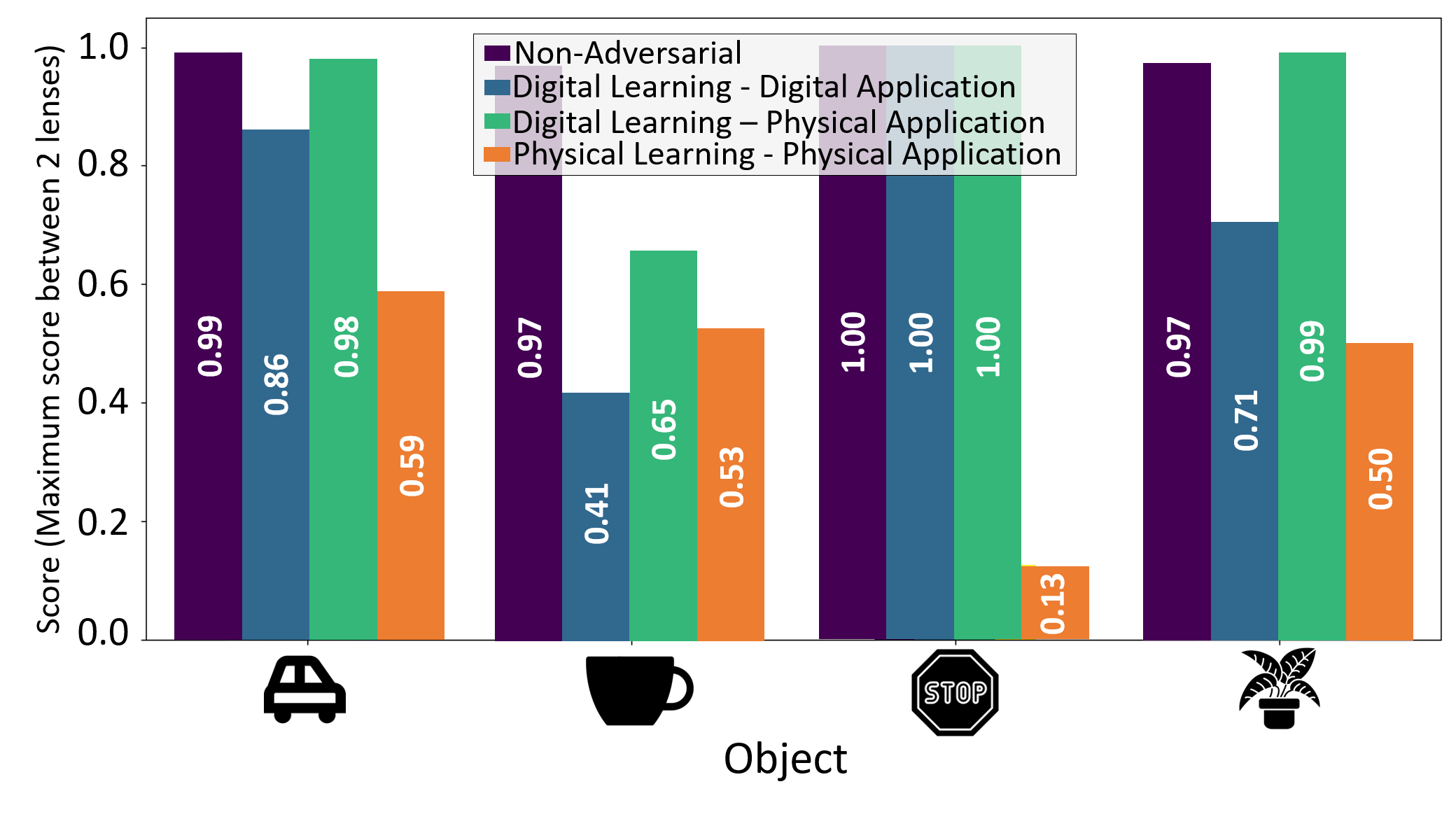}
    \caption{DPatch against YOLOv3 using a stereoscopic camera.}
    \label{fig:dpatch_stereo}
\end{subfigure}
\hfill
\begin{subfigure}[b]{0.49\textwidth}
    \includegraphics[width=\linewidth]{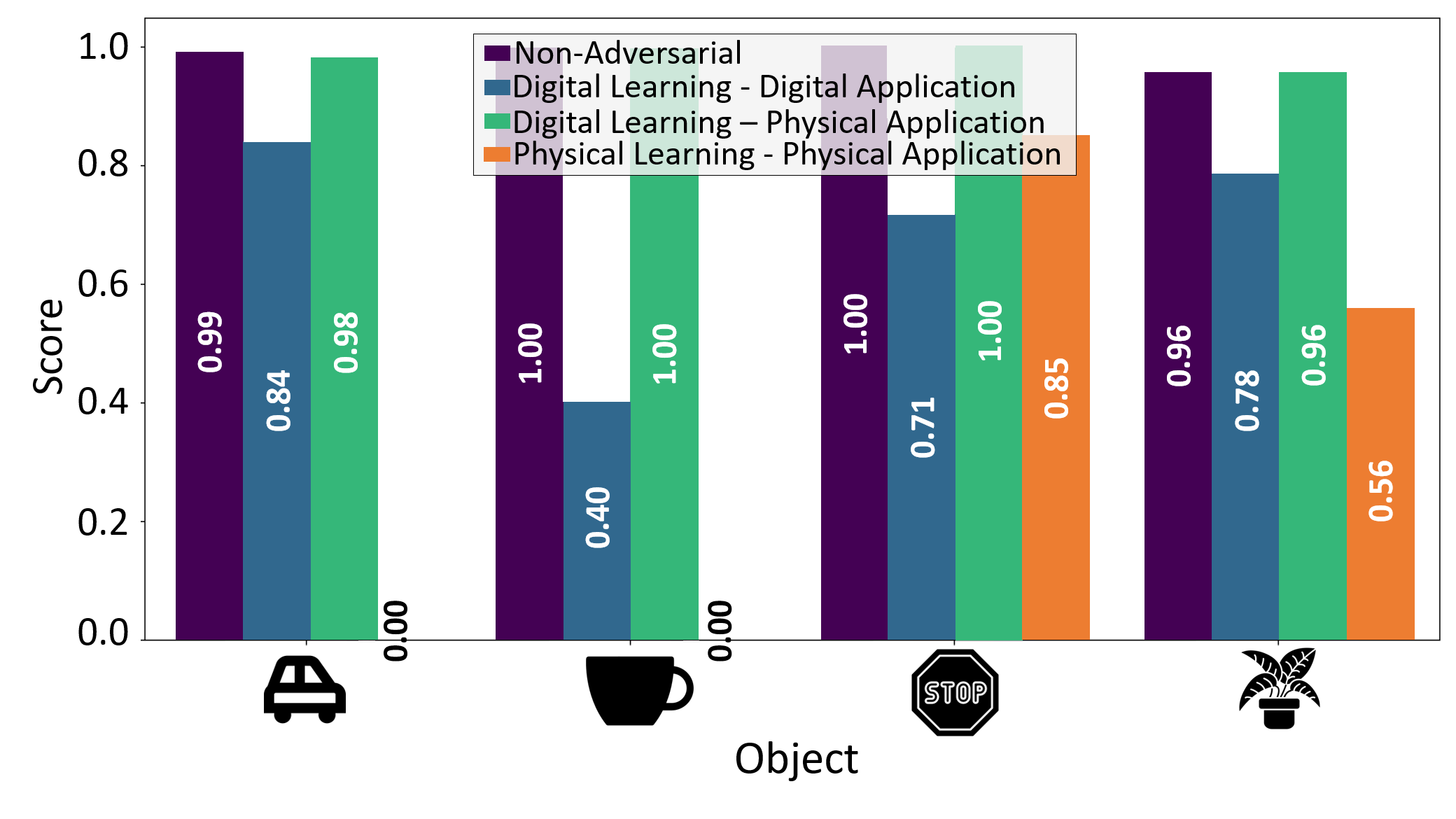}
    \caption{NAP against Faster R-CNN using a monocular camera.}
    \label{fig:nap_monocular}
\end{subfigure}
\hfill
\begin{subfigure}[b]{0.49\textwidth}
    \includegraphics[width=\linewidth]{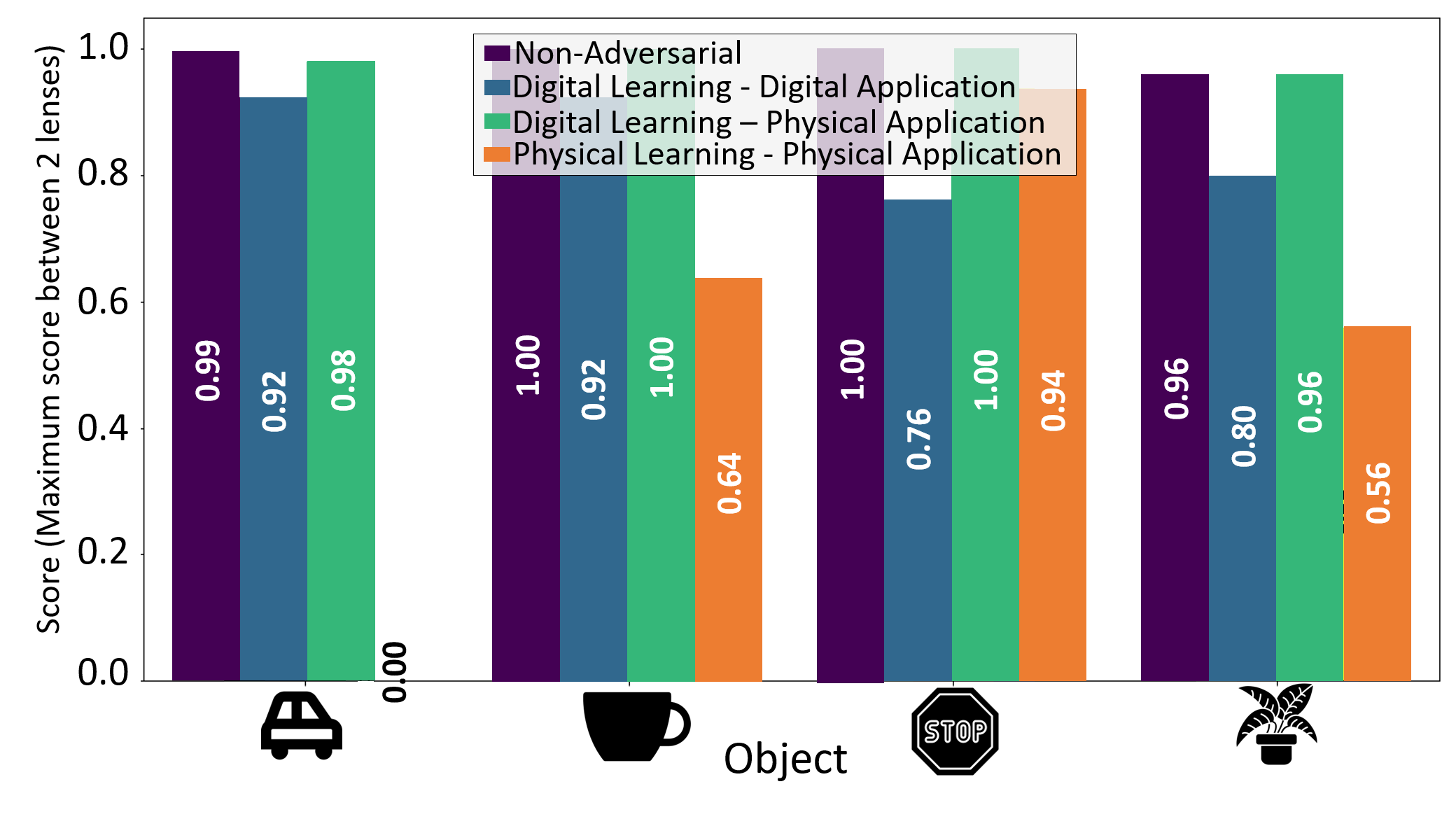}
    \caption{NAP against Faster R-CNN using a stereoscopic camera.}
    \label{fig:nap_stereo}
\end{subfigure}

\caption{Target object confidence scores for different attack and camera setups. 
The purple bar represents the confidence score of the object without a patch (non-adversarial), the dark blue bar represents the confidence score when the patch was learned and applied in the digital domain (DL-DA), the dark green bar represents the confidence score when the patch was learned in the digital domain and applied in the physical domain (DL-PA), and the orange bar represents the confidence score when the patch was learned and applied in the physical domain (PL-PA).}
\label{fig:bars_eval1}
\vspace{-1em}
\end{figure*}

\subsubsection{Results}
The results are presented in Figure \ref{fig:bars_eval1}. For all evaluated target objects (car, stop sign, potted plant, cup), the PL-PA scenario shows a significant reduction in confidence scores, highlighting the effectiveness and potential of utilizing E2E learning in the physical domain.

For the DPatch attack on YOLOv3 (Figures \ref{fig:dpatch_monocular} and \ref{fig:dpatch_stereo}) in the monocular camera setup, the PL-PA scenario reduced the confidence score of the stop sign to 0, the car to 0.16, the potted plant to 0.26, and the cup to 0.15, compared to higher scores for the DL-PA scenario (1.00 for the stop sign, 0.96 for the car, 0.86 for the potted plant, and 0.65 for the cup). Similarly, in the stereoscopic camera setup, the PL-PA scenario reduced confidence scores of the stop sign (0.13), car (0.59), potted plant (0.50), and cup (0.53), compared to 1.00, 0.98, 0.99, and 0.65, respectively, for the DL-PA scenario.

For the NAP attack on Faster R-CNN (Figures~\ref{fig:nap_monocular} and \ref{fig:nap_stereo}) in the monocular camera setup, the PL-PA scenario reduced the confidence scores of the cup and car to 0. The stereoscopic camera setup also produced a confidence reduction; the car’s score was reduced to 0 and the cup's to 0.64. This contrasts with the significantly higher confidence scores for the DL-PA scenario (0.98 for the car and 1.00 for the cup). For both camera setups in the PL-PA scenario, the stop sign confidence scores were slightly higher than the other target objects but remained lower than those in the DL-PA scenario.

In conclusion, the PL-PA scenario, where both patch learning and attack execution occur in the physical domain, is more effective at evading object detection compared to the DL-PA scenario, both in monocular and stereoscopic camera setups.

\begin{insight} \label{insight:eval1_various_attacks}
PAPLA is effective at improving the performance of various adversarial attacks.
\end{insight}
\begin{insight} \label{insight:eval1_various_objects}
PAPLA's PL-PA performance improvement over DL-PA methods is independent of the target object.
\end{insight}

\subsection{Robustness Against Various Object Detectors}
\label{subsec:various_detectors}

Here, we evaluate the robustness of PAPLA's ability to improve the performance of DL-PA adversarial attack methods on various \textit{object detectors}.

\begin{figure} 
	\centering
	\includegraphics[width=0.98\linewidth]{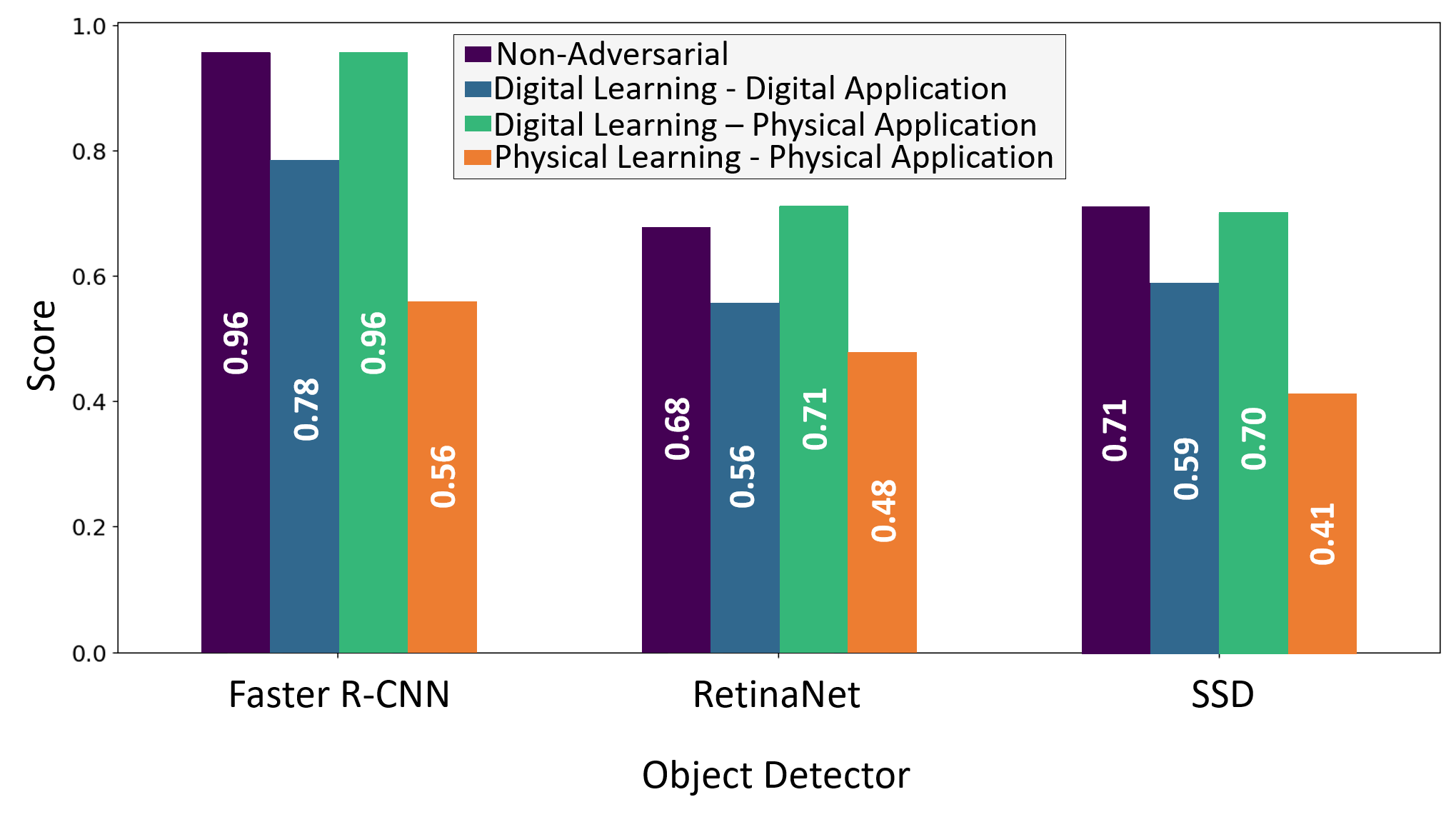}
	\caption{Performance comparison of NAP on a potted plant object against Faster R-CNN, RetinaNet, and SSD, evaluated in four scenarios: non-adversarial, DL-DA, DL-PA, and PL-PA.}
	\label{fig:bars_eval2}
\end{figure}

\subsubsection{Experimental Setup}
We evaluated PAPLA's performance at improving the NAP~\cite{hu2021naturalistic} attack's confidence score reduction on three object detectors: Faster R-CNN \cite{ren2016faster}, RetinaNet \cite{ross2017focal}, and SSD \cite{liu2016ssd}. The target object in this evaluation was a potted plant. We evaluated the performance of the NAP~\cite{hu2021naturalistic} attack under four different scenarios: non-adversarial (a clean potted plant), DL-DA, DL-PA, and PL-PA.
We used the same experimental setup as described in Section~\ref{subsec:various_objs}. The potted plant had a size of \textit{34x20cm}, with a patch size of \textit{4x4cm}, and was captured from a distance of 0.6 meters at a $0\degree$ angle to the camera.

\subsubsection{Results}
The results of this evaluation, as presented in Figure~\ref{fig:bars_eval2}, demonstrate the effectiveness of the NAP attack across different object detectors: Faster R-CNN, RetinaNet, and SSD. 

For Faster R-CNN, in the non-adversarial scenario, the confidence score was 0.96, dropping to 0.78 in the DL-DA scenario. In the DL-PA scenario, the confidence score remained relatively high at 0.96. However, in the PL-PA scenario, the confidence score dropped to 0.56, showing the effectiveness of the PL-PA approach.

For RetinaNet, the confidence score was 0.68 in the non-adversarial scenario, 0.56 in the DL-DA scenario, and 0.71 in the DL-PA scenario. In the PL-PA scenario, the confidence score dropped to 0.48, indicating the robustness of PAPLA's PL-PA confidence reduction.

Lastly, for SSD, the non-adversarial confidence score was 0.71, 0.59 in the DL-DA scenario, and 0.70 in the DL-PA scenario. In the PL-PA scenario, the confidence score dropped to 0.41, a significant reduction compared to the non-adversarial and DL-PA setups.

These results indicate that the PL-PA scenario consistently reduces confidence scores across all evaluated object detectors compared to the DL-PA scenario.

In conclusion, the PL-PA scenario leveraged by PAPLA is capable of improving the confidence reduction of attacks normally performed in the DL-PA scenario across various object detectors.

\begin{insight} \label{insight:eval2_various_detectors}
PAPLA is effective at improving the confidence reduction of DL-PA attacks against various object detectors.
\end{insight}

\subsection{Evaluating Image Quality Using $L_{2}$ and $L_{\infty}$}

Here, we evaluate the image quality of the recorded scenes in each adversarial patch learning scenario: DL-DA, DL-PA, and PL-PA (PAPLA). We use $L_{2}$ and $L_{\infty}$ norms to assess and compare the quality of the results across the three scenarios.
We examine the image quality of the attacks conducted in Sections \ref{subsec:various_objs}, \ref{subsec:various_detectors}, and \ref{subsec:surface_analysis}, using a clean object image without a patch as a baseline for comparison. The experimental setup remains consistent with the settings used in Sections \ref{subsec:surface_analysis}, \ref{subsec:various_objs}, and \ref{subsec:various_detectors}.

\begin{figure} 
	\centering
	\includegraphics[width=0.98\linewidth]{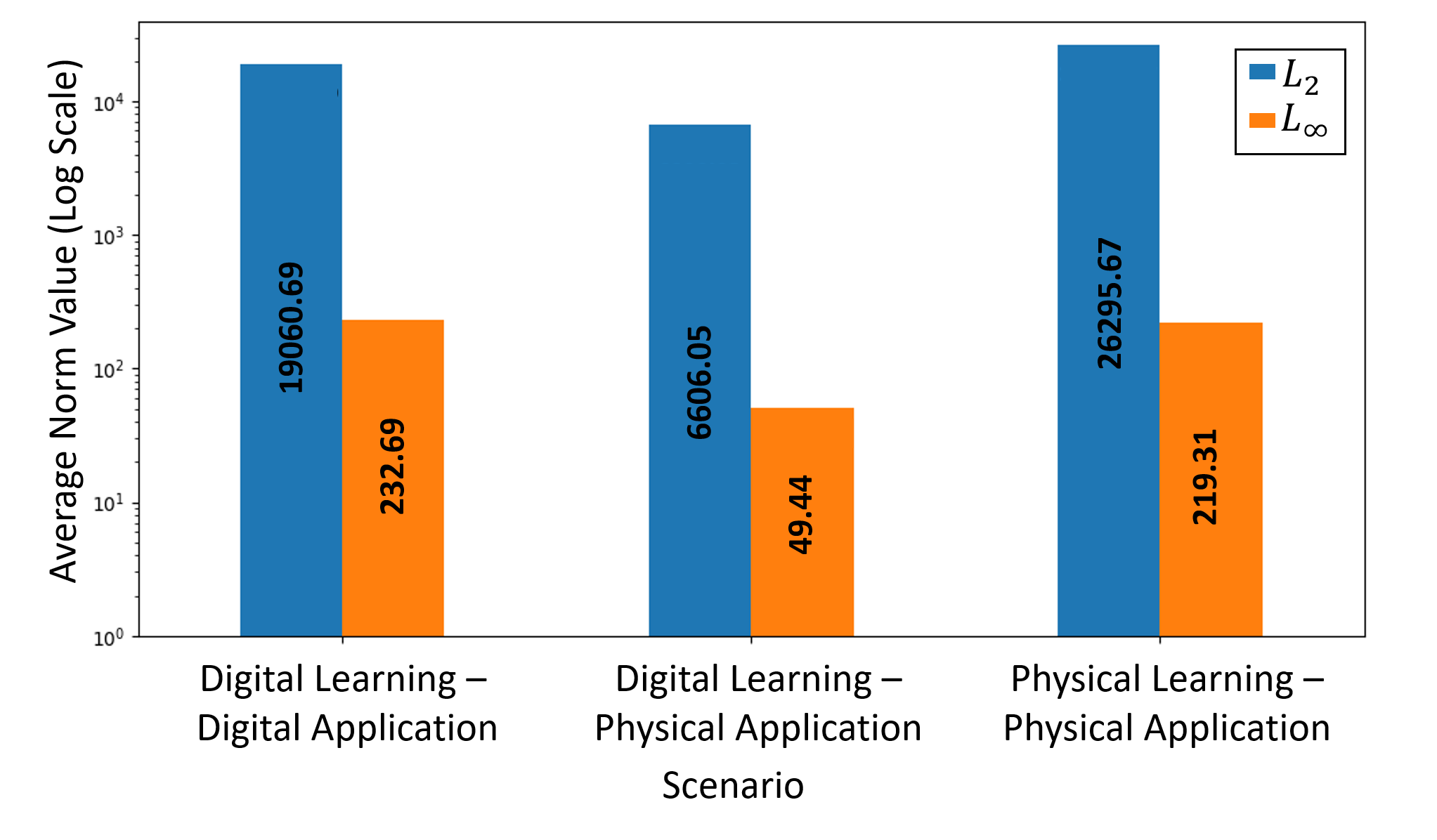}
	\caption{Average $L_{2}$ and $L_{\infty}$ norm values for each scenario (DL-DA, DL-PA, and PL-PA) on a log scale.}
	\label{fig:norms_scenarios}
\end{figure}

\subsubsection{Results}
Figure~\ref{fig:norms_scenarios} presents the image qualities obtained in the three scenarios (DL-DA, DL-PA, and PL-PA) using $L_{2}$ and $L_{\infty}$ norms.

According to the $L_{2}$ norm, the PL-PA scenario (PAPLA) consistently produced the highest norm values, with an average of 26,295.67, indicating that images captured in this scenario had the largest differences compared to the clean object images. This was followed by the DL-DA scenario, with an average $L_{2}$ norm of 19,060.69, and the DL-PA scenario, which produced the lowest average $L_{2}$ norm at 6,606.05.
This suggests that scenes captured in the DL-PA scenario closely resemble the original scene than the other two scenarios. This could be due to utilizing a projector-based patch application in the PL-PA scenario, with light emanation affecting the surrounding scene beyond the areas of the patch.

When examining the $L_{\infty}$ norm, the DL-DA scenario had the highest average value at 232.69, indicating the greatest single-pixel differences between the images and their clean counterparts. The PL-PA scenario had a slightly lower $L_{\infty}$ norm of 219.31, while DL-PA again had the lowest $L_{\infty}$ norm at 49.44. These results further highlight that DL-PA produces the least distortion in terms of both maximum pixel differences and overall image differences.

In conclusion, while PAPLA's PL-PA demonstrates enhanced robustness in reducing detection confidence scores and evading object detection, the lighting emitted by the projector in the physical domain introduces more significant changes to the image. In Section \ref{sec:analysis}, we further investigate the impact of the projector lighting intensity on PAPLA's performance.

\vspace{2mm}
\begin{insight} \label{insight:image_quality}
The DL-PA scenario results in the least image distortion, preserving the image better than both the DL-DA and PL-PA scenarios. PAPLA's PL-PA scenario introduces more significant changes, influenced by factors such as the intensity of the projector's lighting in the physical domain.
\end{insight}

\subsection{Transferability of Patches Across Object Detectors}
Here, we evaluate the transferability of adversarial patches created using PAPLA across different object detectors. Our goal is to examine the effectiveness of patches intended for specific attacks (NAP against Faster R-CNN and DPatch against YOLOv3) when tested on various object detectors, including SSD, RetinaNet, and YOLOv11. We compare the results with those obtained from a DL-DA scenario. This comparison allows us to analyze the effectiveness and transferability of physical patches in real-world settings.

\subsubsection{Experimental Setup}
We created adversarial patches targeting specific object detectors using (1) PAPLA, our PL-PA framework and (2) using the original DL-DA attacks. The patches were generated for the NAP attack against Faster R-CNN and the DPatch attack against YOLOv3, then tested on various object detectors to assess transferability in both PL-PA and DL-DA settings. We evaluated four target objects: a potted plant, a car, a stop sign, and a cup. For each target object, we performed ten runs to calculate the average clean and patched detection confidence scores. We measured the percentage difference in confidence scores between clean and patched objects for each detector. For further details on the experimental setup, see the experimental setup description in Section~\ref{subsubsec:various_objs_exp_setup}.

\begin{table}[]
\centering
\caption{Transferability of Digital Learning - Digital Application and Physical Learning - Physical Application Patches for NAP targeting Faster R-CNN. "-" indicates failure to detect the clean object.}
\resizebox{\columnwidth}{!}{%
\begin{Huge} 
\begin{tabular}{|c|c|cc|}
\hline
\textbf{Target Object} & \textbf{Tested Detector} & \multicolumn{2}{c|}{\textbf{Conf. Diff. (\%)}} \\
\cline{3-4}
                       &                          & \textbf{\makecell{DL-DA}} & \textbf{\makecell{PL-PA}} \\
\hline
\multirow{5}{*}{\textbf{Potted Plant}} & Faster R-CNN & 18.0\% & \textbf{41.6\%} \\
                                       & SSD          & \textbf{13.4\%} & 3.0\% \\
                                       & RetinaNet    & 16.1\% & \textbf{40.4\%} \\
                                       & YOLOv3       & 0\%    & 0\% \\
                                       & YOLOv11      & 40.8\% & \textbf{100.0\%} \\
\cline{1-4}
\multirow{5}{*}{\textbf{Car}}          & Faster R-CNN & 15.4\% & \textbf{100.0\%} \\
                                       & SSD          & 39.2\% & \textbf{100.0\%} \\
                                       & RetinaNet    & 0\%    & \textbf{23.9\%} \\
                                       & YOLOv3       & \textbf{100.0\%} & 47.1\% \\
                                       & YOLOv11      & -      & - \\
\cline{1-4}
\multirow{5}{*}{\textbf{Stop Sign}}    & Faster R-CNN & \textbf{28.5\%} & 14.9\% \\
                                       & SSD          & 45.8\% & \textbf{100.0\%} \\
                                       & RetinaNet    & \textbf{5.2\%} & 0.1\% \\
                                       & YOLOv3       & 0\%    & \textbf{32.7\%} \\
                                       & YOLOv11      & \textbf{38.3\%} & 33.3\% \\
\cline{1-4}
\multirow{5}{*}{\textbf{Cup}}          & Faster R-CNN & 59.6\% & \textbf{100.0\%} \\
                                       & SSD          & 100.0\% & 100.0\% \\
                                       & RetinaNet    & \textbf{100.0\%} & 44.9\% \\
                                       & YOLOv3       & \textbf{22.9\%} & 13.8\% \\
                                       & YOLOv11      & 100.0\% & 100.0\% \\
\hline
\multicolumn{2}{|c|}{\textbf{Average Percentage Confidence Difference}} & 39.1\% & \textbf{52.4\%} \\
\hline
\end{tabular}%
\end{Huge} 
}
\label{table:transferability_nap}
\end{table}

\begin{table}[]
\centering
\caption{Transferability of Digital Learning - Digital Application and Physical Learning - Physical Application Patches for DPatch targeting YOLOv3. "-" indicates failure to detect the clean object.}
\resizebox{\columnwidth}{!}{%
\begin{Huge} 
\begin{tabular}{|c|c|cc|}
\hline
\textbf{Target Object} & \textbf{Tested Detector} & \multicolumn{2}{c|}{\textbf{Conf. Diff. (\%)}} \\
\cline{3-4}
                       &                          & \textbf{\makecell{DL-DA}} & \textbf{\makecell{PL-PA}} \\
\hline
\multirow{5}{*}{\textbf{Potted Plant}} & YOLOv3      & 21.8\% & \textbf{71.0\%} \\
                                       & SSD         & 0\%    & 0\% \\
                                       & RetinaNet   & 6.3\%  & \textbf{19.3\%} \\
                                       & Faster R-CNN & 7.2\%  & \textbf{14.2\%} \\
                                       & YOLOv11     & 32.9\% & \textbf{100.0\%} \\
\cline{1-4}
\multirow{5}{*}{\textbf{Car}}          & YOLOv3      & 16.5\% & \textbf{84.3\%} \\
                                       & SSD         & 0\%    & \textbf{4.3\%} \\
                                       & RetinaNet   & 12.3\% & \textbf{17.2\%} \\
                                       & Faster R-CNN & \textbf{1.3\%} & 0.1\% \\
                                       & YOLOv11     & -      & - \\
\cline{1-4}
\multirow{5}{*}{\textbf{Stop Sign}}    & YOLOv3      & 12.1\% & \textbf{100.0\%} \\
                                       & SSD         & 0\%    & 0\% \\
                                       & RetinaNet   & 0\% & 0\% \\
                                       & Faster R-CNN & 0\% & 0\% \\
                                       & YOLOv11     & 4.5\% & \textbf{31.4\%} \\
\cline{1-4}
\multirow{5}{*}{\textbf{Cup}}          & YOLOv3      & 57.3\% & \textbf{84.1\%} \\
                                       & SSD         & \textbf{100.0\%} & 21.5\% \\
                                       & RetinaNet   & 4.0\%  & \textbf{24.9\%} \\
                                       & Faster R-CNN & 0\%    & \textbf{3.9\%} \\
                                       & YOLOv11     & 100.0\% & 100.0\% \\
\hline
\multicolumn{2}{|c|}{\textbf{Average Percentage Confidence Difference}} & 19.8\% & \textbf{35.6\%} \\
\hline
\end{tabular}%
\end{Huge} 
}
\label{table:transferability_dpatch}
\end{table}

\subsubsection{Results}

The results are summarized in Tables~\ref{table:transferability_nap} and~\ref{table:transferability_dpatch}. These tables highlight the effectiveness of the NAP and DPatch attacks when using PAPLA, comparing the percentage difference in detection confidence between clean and patched objects for both scenarios across various object detectors. 

For patches generated by the \textbf{NAP attack} targeting Faster R-CNN, the average percentage confidence difference was 39.1\% in the DL-DA scenario and 52.4\% in the PL-PA scenario, indicating greater transferability success across different object detectors in the PL-PA scenario. The detailed results show varying effectiveness across different object detectors:
For the \textbf{potted plant}, the NAP patch reduced the confidence score by 41.6\% on Faster R-CNN in the PL-PA scenario, with similar effectiveness on RetinaNet (40.4\%) and a complete reduction (100.0\%) on YOLOv11. The DL-DA scenario showed lower effectiveness, with the highest reduction at 40.8\% on YOLOv11.
For the \textbf{car} object, the PL-PA scenario demonstrated a 100.0\% reduction on Faster R-CNN and SSD, with moderate reductions on RetinaNet (23.9\%) and YOLOv3 (47.1\%). In the DL-DA scenario, the patch achieved a 39.2\% reduction on SSD and a complete confidence reduction (100.0\%) on YOLOv3, indicating a stronger effect on YOLOv3 in DL-DA compared to PL-PA.
The \textbf{stop sign} object showed varied results across object detectors. In the PL-PA scenario, the patch achieved a 100.0\% reduction on SSD and a moderate reduction on YOLOv3 (32.7\%), while producing minimal reduction on RetinaNet (0.1\%) and a 14.9\% reduction on Faster R-CNN. In the DL-DA scenario, SSD had a reduction of 45.8\%, with similar low reductions on RetinaNet (5.2\%) and a slightly stronger effect on Faster R-CNN (28.5\%) compared to PL-PA.
For the \textbf{cup}, the PL-PA patches achieved high confidence reductions on most object detectors, including 100.0\% on Faster R-CNN, SSD, and YOLOv11, while DL-DA also achieved complete confidence reduction on SSD, YOLOv11, and RetinaNet.

For the \textbf{DPatch attack} targeting YOLOv3, the results showed varying effectiveness, with average reductions of 19.8\% in the DL-DA scenario and 35.6\% in the PL-PA scenario. Detailed analysis of each object reveals some notable differences:
For the \textbf{potted plant}, the PL-PA patch was more effective on YOLOv3 (71.0\%) and YOLOv11 (100.0\%), compared to the DL-DA scenario, which had a maximum reduction of 32.9\% on YOLOv11.
The \textbf{car} object showed high effectiveness in the PL-PA scenario on YOLOv3, with an 84.3\% reduction, compared to the DL-DA scenario with a 16.5\% reduction. SSD showed low confidence reductions across both scenarios, indicating its robustness against the DPatch attack for this target object.
For the \textbf{stop sign}, the PL-PA scenario achieved complete confidence reduction on YOLOv3, while the DL-DA scenario had a minimal impact across all object detectors, with a maximum reduction of 12.1\% on YOLOv3.
The \textbf{cup} object displayed high reductions in both scenarios, with 84.1\% reduction on YOLOv3 in the PL-PA scenario and 57.3\% in the DL-DA scenario. YOLOv11 was vulnerable in both scenarios, showing a 100.0\% reduction both times.

\begin{insight} \label{insight:transferability}
While PAPLA's patches tend to have a higher average confidence reduction across object detectors, the results vary per object detector and target object. PL-PA is more effective on YOLOv3 and YOLOv11 for several target objects. However, DL-DA also demonstrates notable confidence reduction in certain cases.
\end{insight}

\subsection{Performance in the Real World}
\label{subsec:outside_env}
Here, we evaluate the performance of the PAPLA framework in real-world outdoor environments. We analyze the learning process and effectiveness of adversarial patches applied to objects under realistic environmental conditions. We perform the NAP \cite{hu2021naturalistic} attack against the YOLOv4 object detector for evaluation.

\subsubsection{Experimental Setup}
\begin{figure}
\centering
\begin{subfigure}[b]{0.44\linewidth}
    \includegraphics[width=\linewidth]{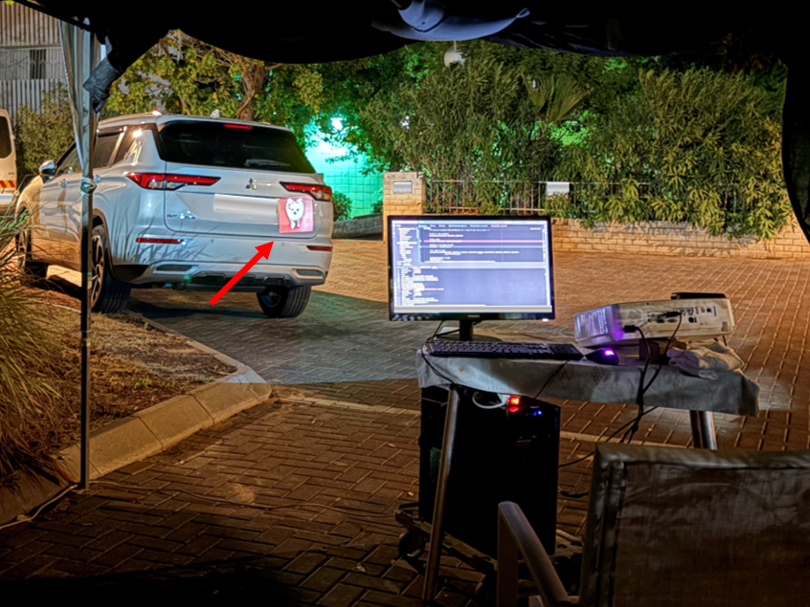}
    \label{fig:car_experimental_setup}
\end{subfigure}
\begin{subfigure}[b]{0.54\linewidth}
    \includegraphics[width=\linewidth]{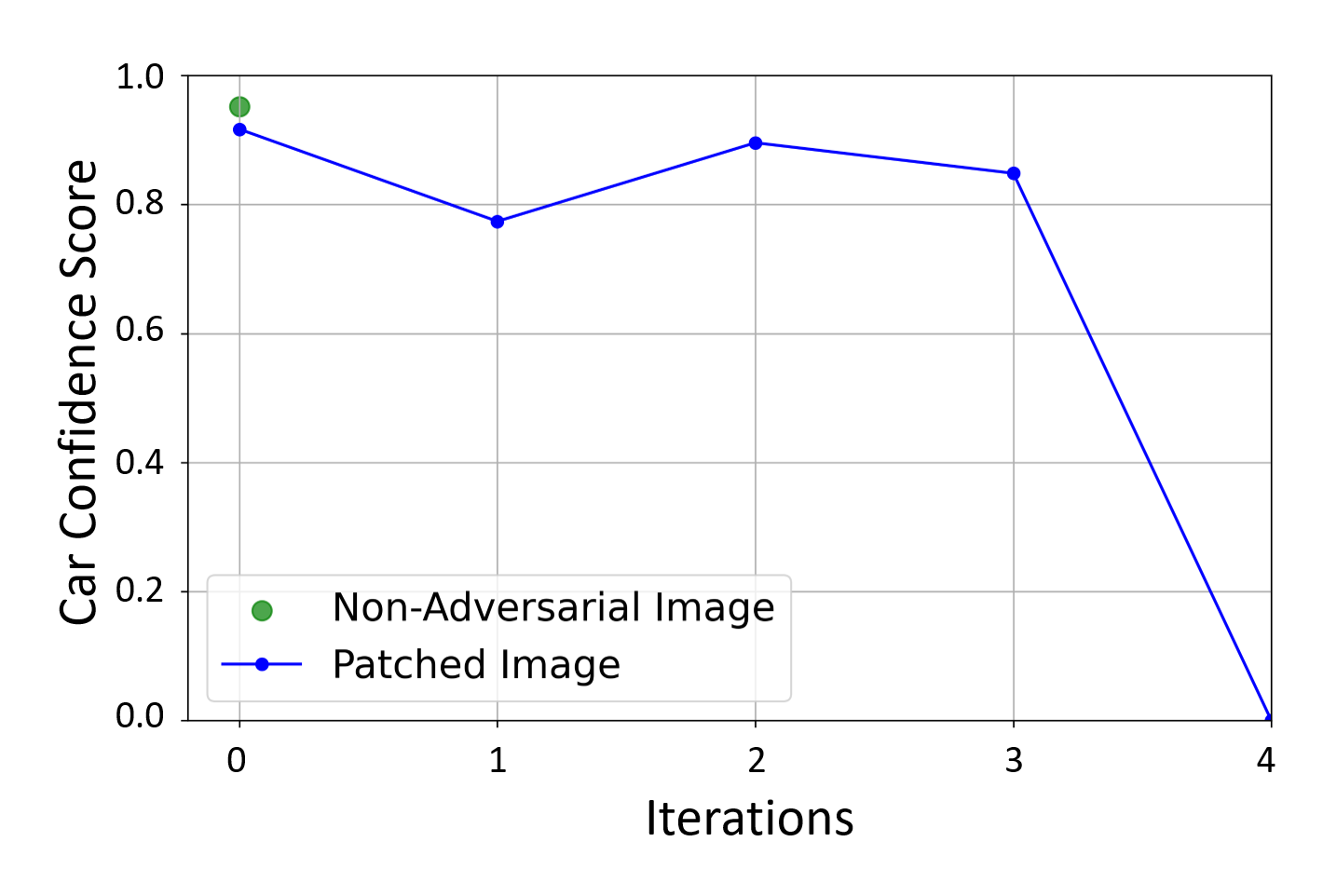}    
    \label{fig:car_outside_learning_process}
\end{subfigure}

\begin{subfigure}[b]{0.44\linewidth}
    \includegraphics[width=\linewidth]{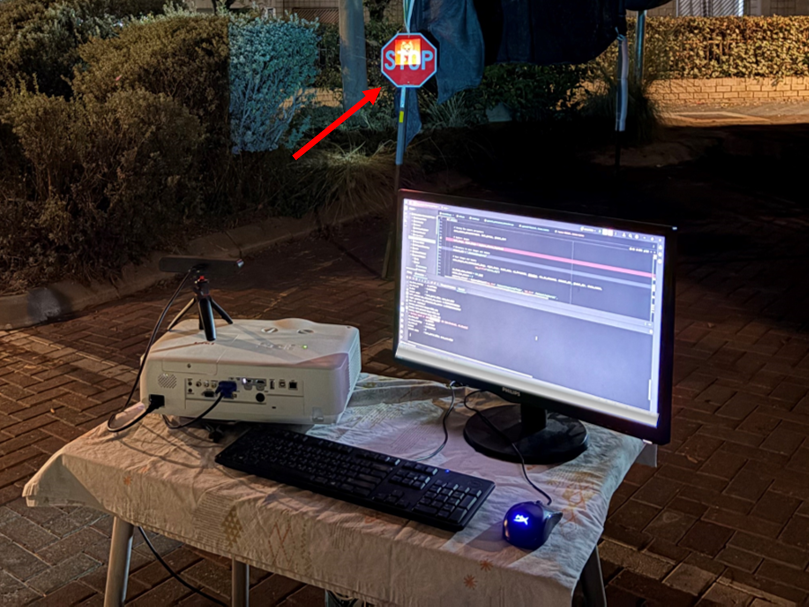}
    \label{fig:stop_sign_experimental_setup}
\end{subfigure}
\begin{subfigure}[b]{0.54\linewidth}
    \includegraphics[width=\linewidth]{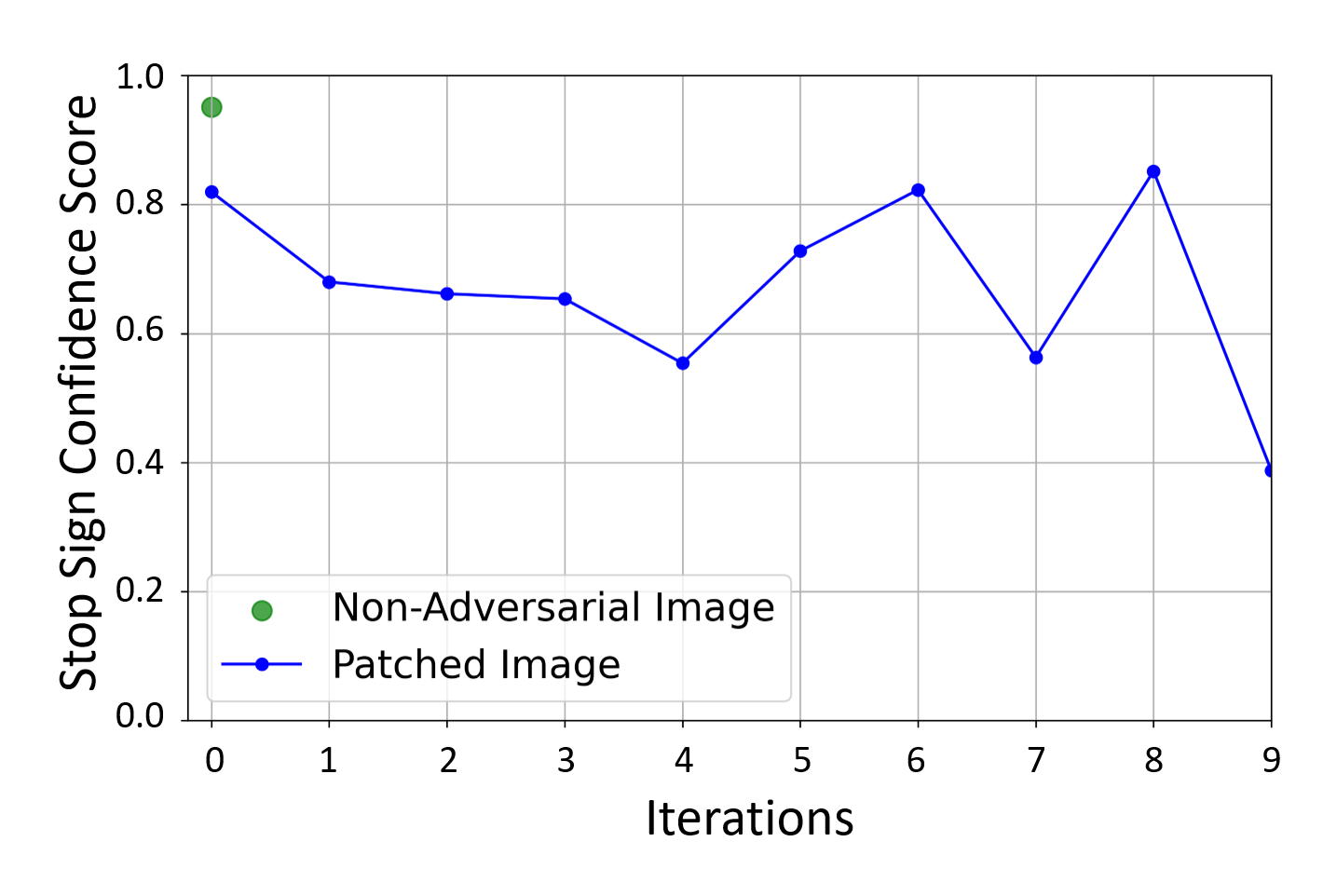}
    \label{fig:stop-sign_outside_learning_process}
\end{subfigure}

\caption{Left: PAPLA setup visualization. Right: Confidence score of a parked car and stop sign targets while conducting PAPLA in an outdoor environment.}
\label{fig:eval5_outside_experimental_setup}
\end{figure}

The experimental setup for the outdoor environment is depicted in Figure~\ref{fig:eval5_outside_experimental_setup}. A projector, camera, and PC were set up outdoors to simulate realistic environmental conditions for patch projection and detection. Two target objects were evaluated: a parked car and a stop sign.

\subsubsection{Results}
The results of the experiments in the outdoor environment are presented in Figure~\ref{fig:eval5_outside_experimental_setup}. The learning process of the adversarial patches, including changes in confidence scores over successive iterations, can also be observed in a demonstration video.\footnote{\url{https://www.youtube.com/watch?v=AtambR-sJD4}}

For the parked car, the confidence score in the non-adversarial scenario was 0.95. By the final iteration, the parked car was not detected at all, with a confidence score of 0. Similarly, for the stop sign, the confidence score started at 0.95 in the non-adversarial scenario and decreased to 0.39 in the last iteration. 

We note the "noisy learning" observed in these experiments. In the outdoor environment, rather than confidence scores that begin at a high value and decrease iteratively, there are fluctuations. This phenomenon is further discussed in Section \ref{sec:limitations}.

\begin{insight} \label{insight:outside_env_effectiveness}
PAPLA demonstrates effectiveness in reducing object detection confidence scores in realistic outdoor scenarios.
\end{insight}
\section{Limitations \& Constraints}
\label{sec:limitations}

While PAPLA can ensure the success of an adversarial attack, it also suffers from the following limitations:

\textbf{(1) Noisy Learning in the Physical Domain.} In Table \ref{tab:image_diff}, we observe significant pixel value changes over short time intervals, even in a controlled and consistent physical domain environment, due to external environmental factors. As a result, the learning process in the physical domain introduces noise, as illustrated in the graphs shown in Figure \ref{fig:eval5_outside_experimental_setup}.

\textbf{(2) Ineffective Against Moving or Hollow Objects.}
PAPLA cannot be used with moving objects, as the projector cannot adjust the patch to match the target's movement in real time. 
Similarly, the method does not work with hollow objects (e.g., bikes, scissors), as such objects do not have a surface which is suitable for projection. Therefore, projecting a patch onto such items is not possible. In such cases, a printed patch has a clear advantage over PAPLA.

\textbf{(3) Performance Affected by Ambient Factors.} As demonstrated in Section \ref{sec:analysis}, PAPLA's performance is affected by various ambient factors, such as lighting conditions, distance, and observation angle. We note that these dependencies are also present in traditional methods that rely either on printed adversarial patches or on projection (e.g., dependency on the angle \cite{lovisotto2021slap,zhang2018camou,sato2024invisible,wen2024opticloak,yufeng2023light}, dependency on the distance \cite{lovisotto2021slap,zhang2018camou,sato2024invisible,wen2024opticloak,yufeng2023light,nassi2020phantom}, dependency on ambient light \cite{lovisotto2021slap,sato2024invisible,wen2024opticloak,yufeng2023light,nassi2020phantom}) and in both cases, must be taken into account to ensure optimal adversarial performance.

\textbf{(4) Constraints of the Threat Model.} PAPLA requires a line-of-sight to the target object to project the adversarial patch during the learning process. 
Additionally, compared to traditional printed patch approaches, PAPLA yields higher $L_2$ and $L_\infty$ norms in the generated patches, resulting in more visually conspicuous patches that are less naturally integrated into the scene. These factors can draw the attention of observers and potentially raise suspicion.

\section{Related Work}
\label{sec:related-work}

Here, we review related work in the domain of physical-domain model evasion attacks, particularly \textit{Patch-based} physical evasion attacks and \textit{Projection-based} physical evasion attacks.

Patch-based physical evasion attacks involve generating a patch in the digital domain, which is then printed and deployed in the physical domain. These attacks pose a significant threat to real-time object detection systems, as they do not require direct access to the deployed target object detector during the attack, only the deployment of the adversarial patch \cite{athalye2018synthesizingrobustadversarialexamples, zolfi2021translucent, guesmi2024dap, huang2023t, zhu2023tpatch, hu2023physically,xu2020adversarial, huang2020universal, wu2020making, hu2021naturalistic, tan2021legitimate, hu2022adversarial, zhu2022infrared, wei2023hotcold, wei2024infrared, chengfull}.
Many methods have been proposed to deploy adversarial patches on shirts \cite{xu2020adversarial, huang2020universal, wu2020making, hu2021naturalistic, tan2021legitimate, hu2022adversarial, zhu2022infrared, wei2023hotcold, wei2024infrared, hu2023physically, guesmi2024dap, chengfull} in order to avoid detection by object detectors.
In addition, physical adversarial patches have been deployed on various items (e.g., hats \cite{komkov2021advhat}, glasses \cite{pautov2019adversarial, hwang2023adversarial, sharif2016accessorize}) or as makeup \cite{zhu2019generating, yin2021adv, lin2022real} in order to evade detection by facial recognition systems.

Projection-based attacks deploy adversarial perturbations using lasers \cite{sato2024invisible,duan2021adversarial}, lights \cite{zhou2018invisiblemaskpracticalattacks,shen2019vla,zhu2021fooling,yufeng2023light,wang2021can}, or projectors \cite{nassi2020phantom,lovisotto2021slap,hu2023adversarial,wen2024opticloak}. 
These methods include projecting virtual objects to deceive advanced driver assistance systems (ADASs) \cite{nassi2020phantom,wen2024opticloak}, using colored light to misclassify objects \cite{hu2023adversarial}, and projecting infrared light to manipulate ADAS perception \cite{wang2021can}. The objectives of these methods range from evading detection by a facial recognition system \cite{zhou2018invisiblemaskpracticalattacks,shen2019vla} to hiding objects from detection by ADASs \cite{sato2024invisible,duan2021adversarial,yufeng2023light,lovisotto2021slap,nassi2020phantom,wang2021can,hu2023adversarial,wen2024opticloak}. 
\section{Conclusion \& Discussion}
\label{sec:discussion}

\textbf{Conclusion \& Discussion.}
This work introduces PAPLA, a novel framework for conducting E2E adversarial learning entirely in the physical domain. 
The findings of this work are not intended to argue against previous works that use traditional digital learning and physical application to apply adversarial attacks in the physical world.
This work intends to demonstrate a new approach that ensures that the application of adversarial attacks using perturbations that were generated in the digital domain will not degrade the performance of the attack when applied in the physical world.

Returning to the research question raised at the beginning: under what constraints might PAPLA yield better results than the traditional digital learning approach? PAPLA is preferred when (1) there is a clear line of sight to the target object for projecting a patch, (2) the object is static, (3) the object has a surface suitable for projection (e.g., not a hollow object like a bicycle), and (4) the ambient conditions are optimal. Examples of possible use cases include autonomous vehicles needing to detect parked cars or road signs. In contrast, when the object is moving, lacks a suitable surface for projection, does not have a clear line of sight for projection, or when the quality of the patch is a priority, traditional digital learning approaches that produce printed patches are preferred.

\textbf{Future Work.}
Future work could explore evaluating PAPLA's effectiveness against existing countermeasure techniques.

\bibliographystyle{IEEEtran}
\bibliography{IEEEabrv,biblio}

\begin{thebibliography}{10}
\providecommand{\url}[1]{#1}
\csname url@samestyle\endcsname
\providecommand{\newblock}{\relax}
\providecommand{\bibinfo}[2]{#2}
\providecommand{\BIBentrySTDinterwordspacing}{\spaceskip=0pt\relax}
\providecommand{\BIBentryALTinterwordstretchfactor}{4}
\providecommand{\BIBentryALTinterwordspacing}{\spaceskip=\fontdimen2\font plus
\BIBentryALTinterwordstretchfactor\fontdimen3\font minus \fontdimen4\font\relax}
\providecommand{\BIBforeignlanguage}[2]{{%
\expandafter\ifx\csname l@#1\endcsname\relax
\typeout{** WARNING: IEEEtran.bst: No hyphenation pattern has been}%
\typeout{** loaded for the language `#1'. Using the pattern for}%
\typeout{** the default language instead.}%
\else
\language=\csname l@#1\endcsname
\fi
#2}}
\providecommand{\BIBdecl}{\relax}
\BIBdecl

\bibitem{ouardirhi2024enhancing}
Z.~Ouardirhi, S.~A. Mahmoudi, and M.~Zbakh, ``Enhancing object detection in smart video surveillance: A survey of occlusion-handling approaches,'' \emph{Electronics}, vol.~13, no.~3, p. 541, 2024.

\bibitem{lubna2021automatic}
Lubna, N.~Mufti, and S.~A.~A. Shah, ``Automatic number plate recognition: A detailed survey of relevant algorithms,'' \emph{Sensors}, vol.~21, no.~9, p. 3028, 2021.

\bibitem{juyal2021deep}
A.~Juyal, S.~Sharma, and P.~Matta, ``Deep learning methods for object detection in autonomous vehicles,'' in \emph{2021 5th International Conference on Trends in Electronics and Informatics (ICOEI)}.\hskip 1em plus 0.5em minus 0.4em\relax IEEE, 2021, pp. 751--755.

\bibitem{hu2021naturalistic}
Y.-C.-T. Hu, B.-H. Kung, D.~S. Tan, J.-C. Chen, K.-L. Hua, and W.-H. Cheng, ``Naturalistic physical adversarial patch for object detectors,'' in \emph{Proceedings of the IEEE/CVF International Conference on Computer Vision}, 2021, pp. 7848--7857.

\bibitem{chen2020hopskipjumpattack}
J.~Chen, M.~I. Jordan, and M.~J. Wainwright, ``Hopskipjumpattack: A query-efficient decision-based attack,'' in \emph{2020 ieee symposium on security and privacy (sp)}.\hskip 1em plus 0.5em minus 0.4em\relax IEEE, 2020, pp. 1277--1294.

\bibitem{guo2019simple}
C.~Guo, J.~Gardner, Y.~You, A.~G. Wilson, and K.~Weinberger, ``Simple black-box adversarial attacks,'' in \emph{International conference on machine learning}.\hskip 1em plus 0.5em minus 0.4em\relax PMLR, 2019, pp. 2484--2493.

\bibitem{brendel2017decision}
W.~Brendel, J.~Rauber, and M.~Bethge, ``Decision-based adversarial attacks: Reliable attacks against black-box machine learning models,'' \emph{arXiv preprint arXiv:1712.04248}, 2017.

\bibitem{athalye2018synthesizingrobustadversarialexamples}
\BIBentryALTinterwordspacing
A.~Athalye, L.~Engstrom, A.~Ilyas, and K.~Kwok, ``Synthesizing robust adversarial examples,'' 2018. [Online]. Available: \url{https://arxiv.org/abs/1707.07397}
\BIBentrySTDinterwordspacing

\bibitem{song2018physical}
D.~Song, K.~Eykholt, I.~Evtimov, E.~Fernandes, B.~Li, A.~Rahmati, F.~Tramer, A.~Prakash, and T.~Kohno, ``Physical adversarial examples for object detectors,'' in \emph{12th USENIX workshop on offensive technologies (WOOT 18)}, 2018.

\bibitem{lee2019physical}
M.~Lee and Z.~Kolter, ``On physical adversarial patches for object detection,'' \emph{arXiv preprint arXiv:1906.11897}, 2019.

\bibitem{eykholt2018robust}
K.~Eykholt, I.~Evtimov, E.~Fernandes, B.~Li, A.~Rahmati, C.~Xiao, A.~Prakash, T.~Kohno, and D.~Song, ``Robust physical-world attacks on deep learning visual classification,'' in \emph{Proceedings of the IEEE conference on computer vision and pattern recognition}, 2018, pp. 1625--1634.

\bibitem{katzav2025adversarialeak}
R.~Katzav, A.~Giloni, E.~Grolman, H.~Saito, T.~Shibata, T.~Omino, M.~Komatsu, Y.~Hanatani, Y.~Elovici, and A.~Shabtai, ``Adversarialeak: External information leakage attack using adversarial samples on face recognition systems,'' in \emph{European Conference on Computer Vision}.\hskip 1em plus 0.5em minus 0.4em\relax Springer, 2025, pp. 288--303.

\bibitem{chen2019shapeshifter}
S.-T. Chen, C.~Cornelius, J.~Martin, and D.~H. Chau, ``Shapeshifter: Robust physical adversarial attack on faster r-cnn object detector,'' in \emph{Machine Learning and Knowledge Discovery in Databases: European Conference, ECML PKDD 2018, Dublin, Ireland, September 10--14, 2018, Proceedings, Part I 18}.\hskip 1em plus 0.5em minus 0.4em\relax Springer, 2019, pp. 52--68.

\bibitem{liu2018dpatch}
X.~Liu, H.~Yang, Z.~Liu, L.~Song, H.~Li, and Y.~Chen, ``Dpatch: An adversarial patch attack on object detectors,'' \emph{arXiv preprint arXiv:1806.02299}, 2018.

\bibitem{zhang2018camou}
Y.~Zhang, H.~Foroosh, P.~David, and B.~Gong, ``Camou: Learning physical vehicle camouflages to adversarially attack detectors in the wild,'' in \emph{International Conference on Learning Representations}, 2018.

\bibitem{thys2019fooling}
S.~Thys, W.~Van~Ranst, and T.~Goedem{\'e}, ``Fooling automated surveillance cameras: adversarial patches to attack person detection,'' in \emph{Proceedings of the IEEE/CVF conference on computer vision and pattern recognition workshops}, 2019, pp. 0--0.

\bibitem{xu2020adversarial}
K.~Xu, G.~Zhang, S.~Liu, Q.~Fan, M.~Sun, H.~Chen, P.-Y. Chen, Y.~Wang, and X.~Lin, ``Adversarial t-shirt! evading person detectors in a physical world,'' in \emph{Computer Vision--ECCV 2020: 16th European Conference, Glasgow, UK, August 23--28, 2020, Proceedings, Part V 16}.\hskip 1em plus 0.5em minus 0.4em\relax Springer, 2020, pp. 665--681.

\bibitem{huang2020universal}
L.~Huang, C.~Gao, Y.~Zhou, C.~Xie, A.~L. Yuille, C.~Zou, and N.~Liu, ``Universal physical camouflage attacks on object detectors,'' in \emph{Proceedings of the IEEE/CVF conference on computer vision and pattern recognition}, 2020, pp. 720--729.

\bibitem{wu2020making}
Z.~Wu, S.-N. Lim, L.~S. Davis, and T.~Goldstein, ``Making an invisibility cloak: Real world adversarial attacks on object detectors,'' in \emph{Computer Vision--ECCV 2020: 16th European Conference, Glasgow, UK, August 23--28, 2020, Proceedings, Part IV 16}.\hskip 1em plus 0.5em minus 0.4em\relax Springer, 2020, pp. 1--17.

\bibitem{zolfi2021translucent}
A.~Zolfi, M.~Kravchik, Y.~Elovici, and A.~Shabtai, ``The translucent patch: A physical and universal attack on object detectors,'' in \emph{Proceedings of the IEEE/CVF conference on computer vision and pattern recognition}, 2021, pp. 15\,232--15\,241.

\bibitem{jing2021too}
P.~Jing, Q.~Tang, Y.~Du, L.~Xue, X.~Luo, T.~Wang, S.~Nie, and S.~Wu, ``Too good to be safe: Tricking lane detection in autonomous driving with crafted perturbations,'' in \emph{30th USENIX Security Symposium (USENIX Security 21)}, 2021, pp. 3237--3254.

\bibitem{tan2021legitimate}
J.~Tan, N.~Ji, H.~Xie, and X.~Xiang, ``Legitimate adversarial patches: Evading human eyes and detection models in the physical world,'' in \emph{Proceedings of the 29th ACM international conference on multimedia}, 2021, pp. 5307--5315.

\bibitem{suryanto2022dta}
N.~Suryanto, Y.~Kim, H.~Kang, H.~T. Larasati, Y.~Yun, T.-T.-H. Le, H.~Yang, S.-Y. Oh, and H.~Kim, ``Dta: Physical camouflage attacks using differentiable transformation network,'' in \emph{Proceedings of the IEEE/CVF Conference on Computer Vision and Pattern Recognition}, 2022, pp. 15\,305--15\,314.

\bibitem{hu2022adversarial}
Z.~Hu, S.~Huang, X.~Zhu, F.~Sun, B.~Zhang, and X.~Hu, ``Adversarial texture for fooling person detectors in the physical world,'' in \emph{Proceedings of the IEEE/CVF conference on computer vision and pattern recognition}, 2022, pp. 13\,307--13\,316.

\bibitem{biton2023adversarial}
D.~Biton, A.~Misra, E.~Levy, J.~Kotak, R.~Bitton, R.~Schuster, N.~Papernot, Y.~Elovici, and B.~Nassi, ``The adversarial implications of variable-time inference,'' in \emph{Proceedings of the 16th ACM Workshop on Artificial Intelligence and Security}, 2023, pp. 103--114.

\bibitem{jia2022fooling}
W.~Jia, Z.~Lu, H.~Zhang, Z.~Liu, J.~Wang, and G.~Qu, ``Fooling the eyes of autonomous vehicles: Robust physical adversarial examples against traffic sign recognition systems,'' \emph{arXiv preprint arXiv:2201.06192}, 2022.

\bibitem{huang2023t}
H.~Huang, Z.~Chen, H.~Chen, Y.~Wang, and K.~Zhang, ``T-sea: Transfer-based self-ensemble attack on object detection,'' in \emph{Proceedings of the IEEE/CVF Conference on Computer Vision and Pattern Recognition}, 2023, pp. 20\,514--20\,523.

\bibitem{zhu2023tpatch}
W.~Zhu, X.~Ji, Y.~Cheng, S.~Zhang, and W.~Xu, ``$\{$TPatch$\}$: A triggered physical adversarial patch,'' in \emph{32nd USENIX Security Symposium (USENIX Security 23)}, 2023, pp. 661--678.

\bibitem{hu2023physically}
Z.~Hu, W.~Chu, X.~Zhu, H.~Zhang, B.~Zhang, and X.~Hu, ``Physically realizable natural-looking clothing textures evade person detectors via 3d modeling,'' in \emph{Proceedings of the IEEE/CVF Conference on Computer Vision and Pattern Recognition}, 2023, pp. 16\,975--16\,984.

\bibitem{guesmi2024dap}
A.~Guesmi, R.~Ding, M.~A. Hanif, I.~Alouani, and M.~Shafique, ``Dap: A dynamic adversarial patch for evading person detectors,'' in \emph{Proceedings of the IEEE/CVF Conference on Computer Vision and Pattern Recognition}, 2024, pp. 24\,595--24\,604.

\bibitem{weirevisiting}
H.~Wei, Z.~Wang, K.~Zhang, J.~Hou, Y.~Liu, H.~Tang, and Z.~Wang, ``Revisiting adversarial patches for designing camera-agnostic attacks against person detection,'' in \emph{The Thirty-eighth Annual Conference on Neural Information Processing Systems}.

\bibitem{chengfull}
Z.~Cheng, Z.~Hu, Y.~Liu, J.~Li, H.~Su, and X.~Hu, ``Full-distance evasion of pedestrian detectors in the physical world,'' in \emph{The Thirty-eighth Annual Conference on Neural Information Processing Systems}.

\bibitem{zhu2022infrared}
X.~Zhu, Z.~Hu, S.~Huang, J.~Li, and X.~Hu, ``Infrared invisible clothing: Hiding from infrared detectors at multiple angles in real world,'' in \emph{Proceedings of the IEEE/CVF Conference on Computer Vision and Pattern Recognition}, 2022, pp. 13\,317--13\,326.

\bibitem{wei2023hotcold}
H.~Wei, Z.~Wang, X.~Jia, Y.~Zheng, H.~Tang, S.~Satoh, and Z.~Wang, ``Hotcold block: Fooling thermal infrared detectors with a novel wearable design,'' in \emph{Proceedings of the AAAI conference on artificial intelligence}, vol.~37, no.~12, 2023, pp. 15\,233--15\,241.

\bibitem{wei2024infrared}
X.~Wei, J.~Yu, and Y.~Huang, ``Infrared adversarial patches with learnable shapes and locations in the physical world,'' \emph{International Journal of Computer Vision}, vol. 132, no.~6, pp. 1928--1944, 2024.

\bibitem{zhu2024infrared}
X.~Zhu, Y.~Liu, Z.~Hu, J.~Li, and X.~Hu, ``Infrared adversarial car stickers,'' in \emph{Proceedings of the IEEE/CVF Conference on Computer Vision and Pattern Recognition}, 2024, pp. 24\,284--24\,293.

\bibitem{lovisotto2021slap}
G.~Lovisotto, H.~Turner, I.~Sluganovic, M.~Strohmeier, and I.~Martinovic, ``$\{$SLAP$\}$: Improving physical adversarial examples with $\{$Short-Lived$\}$ adversarial perturbations,'' in \emph{30th USENIX Security Symposium (USENIX Security 21)}, 2021, pp. 1865--1882.

\bibitem{wen2024opticloak}
H.~Wen, S.~Chang, L.~Zhou, W.~Liu, and H.~Zhu, ``Opticloak: Blinding vision-based autonomous driving systems through adversarial optical projection,'' \emph{IEEE Internet of Things Journal}, 2024.

\bibitem{hu2023adversarial}
C.~Hu, W.~Shi, and L.~Tian, ``Adversarial color projection: A projector-based physical-world attack to dnns,'' \emph{Image and Vision Computing}, vol. 140, p. 104861, 2023.

\bibitem{hwang2023adversarial}
R.-H. Hwang, J.-Y. Lin, S.-Y. Hsieh, H.-Y. Lin, and C.-L. Lin, ``Adversarial patch attacks on deep-learning-based face recognition systems using generative adversarial networks,'' \emph{Sensors}, vol.~23, no.~2, p. 853, 2023.

\bibitem{komkov2021advhat}
S.~Komkov and A.~Petiushko, ``Advhat: Real-world adversarial attack on arcface face id system,'' in \emph{2020 25th international conference on pattern recognition (ICPR)}.\hskip 1em plus 0.5em minus 0.4em\relax IEEE, 2021, pp. 819--826.

\bibitem{lin2022real}
C.-S. Lin, C.-Y. Hsu, P.-Y. Chen, and C.-M. Yu, ``Real-world adversarial examples via makeup,'' in \emph{ICASSP 2022-2022 IEEE International Conference on Acoustics, Speech and Signal Processing (ICASSP)}.\hskip 1em plus 0.5em minus 0.4em\relax IEEE, 2022, pp. 2854--2858.

\bibitem{wei2023unified}
X.~Wei, Y.~Huang, Y.~Sun, and J.~Yu, ``Unified adversarial patch for cross-modal attacks in the physical world,'' in \emph{Proceedings of the IEEE/CVF International Conference on Computer Vision}, 2023, pp. 4445--4454.

\bibitem{nassi2020phantom}
B.~Nassi, Y.~Mirsky, D.~Nassi, R.~Ben-Netanel, O.~Drokin, and Y.~Elovici, ``Phantom of the adas: Securing advanced driver-assistance systems from split-second phantom attacks,'' in \emph{Proceedings of the 2020 ACM SIGSAC conference on computer and communications security}, 2020, pp. 293--308.

\bibitem{nicolae2018adversarial}
M.-I. Nicolae, M.~Sinn, M.~N. Tran, B.~Buesser, A.~Rawat, M.~Wistuba, V.~Zantedeschi, N.~Baracaldo, B.~Chen, H.~Ludwig \emph{et~al.}, ``Adversarial robustness toolbox v1. 0.0,'' \emph{arXiv preprint arXiv:1807.01069}, 2018.

\bibitem{redmon2018yolov3}
J.~Redmon and A.~Farhadi, ``Yolov3: An incremental improvement,'' \emph{arXiv preprint arXiv:1804.02767}, 2018.

\bibitem{ren2016faster}
S.~Ren, K.~He, R.~Girshick, and J.~Sun, ``Faster r-cnn: Towards real-time object detection with region proposal networks,'' \emph{IEEE transactions on pattern analysis and machine intelligence}, vol.~39, no.~6, pp. 1137--1149, 2016.

\bibitem{ross2017focal}
T.-Y. Ross and G.~Doll{\'a}r, ``Focal loss for dense object detection,'' in \emph{proceedings of the IEEE conference on computer vision and pattern recognition}, 2017, pp. 2980--2988.

\bibitem{liu2016ssd}
W.~Liu, D.~Anguelov, D.~Erhan, C.~Szegedy, S.~Reed, C.-Y. Fu, and A.~C. Berg, ``Ssd: Single shot multibox detector,'' in \emph{Computer Vision--ECCV 2016: 14th European Conference, Amsterdam, The Netherlands, October 11--14, 2016, Proceedings, Part I 14}.\hskip 1em plus 0.5em minus 0.4em\relax Springer, 2016, pp. 21--37.

\bibitem{sato2024invisible}
T.~Sato, S.~H.~V. Bhupathiraju, M.~Clifford, T.~Sugawara, Q.~A. Chen, and S.~Rampazzi, ``Invisible reflections: Leveraging infrared laser reflections to target traffic sign perception,'' \emph{arXiv preprint arXiv:2401.03582}, 2024.

\bibitem{yufeng2023light}
L.~Yufeng, Y.~Fengyu, L.~Qi, L.~Jiangtao, and C.~Chenhong, ``Light can be dangerous: Stealthy and effective physical-world adversarial attack by spot light,'' \emph{Computers \& Security}, vol. 132, p. 103345, 2023.

\bibitem{pautov2019adversarial}
M.~Pautov, G.~Melnikov, E.~Kaziakhmedov, K.~Kireev, and A.~Petiushko, ``On adversarial patches: real-world attack on arcface-100 face recognition system,'' in \emph{2019 International Multi-Conference on Engineering, Computer and Information Sciences (SIBIRCON)}.\hskip 1em plus 0.5em minus 0.4em\relax IEEE, 2019, pp. 0391--0396.

\bibitem{sharif2016accessorize}
M.~Sharif, S.~Bhagavatula, L.~Bauer, and M.~K. Reiter, ``Accessorize to a crime: Real and stealthy attacks on state-of-the-art face recognition,'' in \emph{Proceedings of the 2016 acm sigsac conference on computer and communications security}, 2016, pp. 1528--1540.

\bibitem{zhu2019generating}
Z.-A. Zhu, Y.-Z. Lu, and C.-K. Chiang, ``Generating adversarial examples by makeup attacks on face recognition,'' in \emph{2019 IEEE International Conference on Image Processing (ICIP)}.\hskip 1em plus 0.5em minus 0.4em\relax IEEE, 2019, pp. 2516--2520.

\bibitem{yin2021adv}
B.~Yin, W.~Wang, T.~Yao, J.~Guo, Z.~Kong, S.~Ding, J.~Li, and C.~Liu, ``Adv-makeup: A new imperceptible and transferable attack on face recognition,'' \emph{arXiv preprint arXiv:2105.03162}, 2021.

\bibitem{duan2021adversarial}
R.~Duan, X.~Mao, A.~K. Qin, Y.~Chen, S.~Ye, Y.~He, and Y.~Yang, ``Adversarial laser beam: Effective physical-world attack to dnns in a blink,'' in \emph{Proceedings of the IEEE/CVF Conference on Computer Vision and Pattern Recognition}, 2021, pp. 16\,062--16\,071.

\bibitem{zhou2018invisiblemaskpracticalattacks}
\BIBentryALTinterwordspacing
Z.~Zhou, D.~Tang, X.~Wang, W.~Han, X.~Liu, and K.~Zhang, ``Invisible mask: Practical attacks on face recognition with infrared,'' 2018. [Online]. Available: \url{https://arxiv.org/abs/1803.04683}
\BIBentrySTDinterwordspacing

\bibitem{shen2019vla}
M.~Shen, Z.~Liao, L.~Zhu, K.~Xu, and X.~Du, ``Vla: A practical visible light-based attack on face recognition systems in physical world,'' \emph{Proceedings of the ACM on Interactive, Mobile, Wearable and Ubiquitous Technologies}, vol.~3, no.~3, pp. 1--19, 2019.

\bibitem{zhu2021fooling}
X.~Zhu, X.~Li, J.~Li, Z.~Wang, and X.~Hu, ``Fooling thermal infrared pedestrian detectors in real world using small bulbs,'' in \emph{Proceedings of the AAAI conference on artificial intelligence}, vol.~35, no.~4, 2021, pp. 3616--3624.

\bibitem{wang2021can}
W.~Wang, Y.~Yao, X.~Liu, X.~Li, P.~Hao, and T.~Zhu, ``I can see the light: Attacks on autonomous vehicles using invisible lights,'' in \emph{Proceedings of the 2021 ACM SIGSAC Conference on Computer and Communications Security}, 2021, pp. 1930--1944.

\end{thebibliography}

\end{document}